\documentclass[11pt,a4paper,english]{article}

\newcommand{\comment}[1]{}

\usepackage{array}
\newcolumntype{C}[1]{>{\centering\arraybackslash$}p{#1}<{$}}

\usepackage{bm}
\usepackage{amsmath}
\usepackage{color, colortbl}
\usepackage{multirow}
\usepackage[usenames,dvipsnames]{xcolor}
\usepackage{slashbox}
\usepackage{xspace}
\usepackage{soul}
\usepackage{algorithm}
\usepackage{algorithmic}
\usepackage{tabularx}

\usepackage{xfrac}
\usepackage{appendix}
\usepackage{graphicx}
\usepackage{array}
\usepackage{caption}
\usepackage{xspace}
\usepackage{amssymb}  

\usepackage{color}
\definecolor{orange}{RGB}{255,127,0}
\definecolor{brown}{RGB}{150,70,0}
\definecolor{green}{RGB}{127,255,127}
\definecolor{darkgreen}{RGB}{0,127,0}
\definecolor{blue}{RGB}{127,127,255}
\definecolor{lightblue}{RGB}{150,150,255}
\definecolor{darkblue}{RGB}{0,0,127}
\definecolor{red}{RGB}{255,90,90}
\definecolor{grey}{RGB}{127,127,127}
\definecolor{pink}{RGB}{255,180,180}

\newcommand{\blue}[1][blue]{\textcolor{blue}{{#1}}}

\newcommand{\green}[1][green]{\textcolor{green}{{#1}}}

\newcommand{\violet}[1][violet]{\textcolor{violet}{{#1}}}
\newcommand{\ra}{\rightarrow}

\newcommand{\gerl}{\violet[\sffamily{gErl}]\xspace}

\usepackage{graphicx}

\usepackage[colorlinks,pagebackref,linktocpage]{hyperref}  

\usepackage{amsthm}

\usepackage[affil-it]{authblk}

\theoremstyle{definition}
\newtheorem{exmp}{Example}[section]

\title{On the definition of a general learning system with user-defined operators
}

\author{Fernando Mart{\mbox{\'{\i}}}nez-Plumed \\ C\`{e}sar Ferri \\ Jos$\acute{\mbox{e}}$ Hern$\acute{\mbox{a}}$ndez-Orallo \\ Mar${\mbox{\'{\i}}}$a Jos$\acute{\mbox{e}}$ Ram${\mbox{\'{\i}}}$rez-Quintana\\  
  {\ttfamily{\{fmartinez,cferri,jorallo,mramirez\}@dsic.upv.es}}}
\affil{DSIC, Universitat Polit\`ecnica de Val\`encia, \\ Cam\'{\i} de Vera s/n, 46022 Val\`encia, Spain.}

\date{\today}

\begin{document}

\maketitle

{
\abstract  In this paper, we push forward the idea of machine learning systems whose operators can be modified and finetuned for each problem. This allows us to propose a learning paradigm where users can write (or adapt) their operators, according to the problem, data representation and the way the information should be navigated. To achieve this goal, data instances, background knowledge, rules, programs and \emph{operators} are all written in the same functional language, Erlang. Since changing operators affect how the search space needs to be explored, heuristics are learnt as a result of a decision process based on reinforcement learning where each action is defined as a choice of operator and rule. As a result, the architecture can be seen as a `system for writing machine learning systems' or to explore new operators where the policy reuse (as a kind of transfer learning) is allowed. States and actions are represented in a $Q$ matrix which is actually a table, from which a supervised model is learnt. This makes it possible to have a more flexible mapping between old and new problems, since we work with an abstraction of rules and actions. We include some examples sharing reuse and the application of the system \gerl to IQ problems.

In order to evaluate \gerl, we will test it against some structured problems: a selection of IQ test tasks and some experiments on some structured prediction problems (list patterns).
\\

}

\section{Introduction}\label{intro}

The number and performance of machine learning techniques dealing with complex, structured data have considerably increased in the past decades. 
However, the performance of these systems is usually linked to a transformation of the feature space (possibly including the outputs as well) to a more convenient, flat, representation, which typically leads to incomprehensible patterns in terms of the transformed (hyper-)space. Alternatively, other approaches do stick to the original problem representation but rely on specialised systems with embedded operators that are only able to deal with specific types of data.

Despite all these approaches and the vindication of more general frameworks for data mining \cite{Dzeroski:2006:TGF:1777194.1777213}, there is no general-purpose machine learning system which can deal with {\em all} of these problems {\em preserving} the problem representation. There are of course several paradigms using, e.g., distances or kernel methods for structured data \cite{Gaertner05,vicentsim06} which can be applied to virtually any kind of data. However, this generality comes at the cost of losing the original problem representation and typically losing the recursive character of many data structures.

Other paradigms, such as inductive programming (ILP \cite{Mug99}, IFP \cite{Kitzelmann10} or IFLP \cite{Ferri-RamirezHR01}), are able to tackle any kind of data thanks to the expressive power of first-order logic (or term rewriting systems). However, each system has a predefined set of operators (e.g., \emph{lgg} \cite{Plo70}, inverse entailment \cite{mug95}, splitting conditions in a decision tree, or others) and an embedded heuristic. Even with the help of background knowledge it is still virtually impossible to deal with, e.g., an XML document, if we do not have the appropriate operators to delve into its structure and an appropriate heuristic to prioritise their application.

In this paper we present and explore a general rule-based learning system \gerl where operators can be defined and customised for each kind of problem. While one particular problem may require generalisation operators, another problem may require operators which add recursive transformations to explore the structure of the data. A right choice of operators can embed transformations on the data but can also determine the way in which rules are generated and transformed, so leading to (apparently) different learning systems. Making the user or the problem adapt its own operators is significantly different to the use of feature transformations or specific background knowledge. In fact, it is also significantly more difficult, since operators can be very complex things and usually embed the essence of a machine learning system. A very simple operator, such as $lgg$, requires several lines of code in almost any programming language, if not more. Writing and adapting a system to a new operator is not always an easy task. As a result, having a system which can work with different kinds of operators at the same time is a challenging proposal beyond the frontiers of the state of the art in machine learning.

In addition, machine learning operators are tools to explore the hypothesis space. Consequently, some operators are usually associated to some heuristic strategies (e.g., generalisation operators and bottom-up strategies). By giving more freedom to the kind of operators a system can use, we lose the capacity to analyse and define particular heuristics to tame the search space. This means that heuristics must be overhauled, in terms of {\em decisions} about the operator that must be used at each particular state of the learning process.

We therefore propose a learning system where operators can be written or modified by the user. Since operators are defined as functions which transform patterns, we clearly need a language for defining operators which can integrate the representation of the examples, patterns and operators. We will argue that functional programming languages, with reflection and higher-order primitives, are appropriate for this, and we will choose a powerful and relatively popular programming language in this family, Erlang \cite{Armstrong07}. A not less important reason for using a functional language is that operators can be understood by the users and properly linked with the data structures used in the examples and background knowledge, so making the specification of new operators easier.
The language also sets the general representation of examples as equations, patterns as rules and models as sets of rules.

From here, we devise a flexible architecture which works with populations of rules and programs, which {\em evolve} as in an evolutionary programming setting or a learning classifier system \cite{Holland00}. Operators are applied to rules and generate new rules, which are combined with existing or new programs. With appropriate operators and using some optimality criteria (based on coverage and simplicity) we will eventually find some good solutions to the learning problem. However, without heuristics, the number of required iterations gets astronomically high. This issue is addressed with a reinforcement learning (RL) approach, where the application of an operator over a rule is seen as a decision problem, for which learning also takes place, guided by the optimality criteria which feed a rewarding module. As a result, the architecture can be seen as a `system for writing machine learning systems' or to explore new operators.

Interestingly, different problems using the same operators can reuse the heuristics. 
Since the RL system determines which rules and operators are used and how they are combined, RL policies can be reused between similar (or totally different) tasks. The knowledge transferred between tasks can be viewed as a bias in the learning of the target task using the information learned in the source task. In order to do that, we use an appropriate abstract feature space for describing the kinds of rules and operators that are giving good solutions (and high rewards), so this history is reused for other problems, even when the task and  operators are different.

The paper is organised as follows. Section \ref{previous} makes a short account of the many approaches and ideas which are related to this proposal. Section \ref{system} introduces \gerl and how operators are expressed and applied. Section \ref{heuristic} describes the RL-based heuristics used to guide the learning process. Section \ref{sec:reuse} describes how \gerl is able to transfer knowledge between tasks.  Section \ref{IQ} includes some experiments  which illustrate how \gerl solves several to IQ problems. Finally, a more comprehensive and thorough analysis is discussed in section \ref{discussion}, which, together with the the section \ref{conclusions}, closes the paper.

\section{Previous work}\label{previous}

Since we propose a general learning system, it is necessarily related to different areas of machine learning such as learning from complex data,  inductive programming, reinforcement learning (RL), Learning Classifiers Systems (LCS), evolutionary techniques, meta-learning, etc., and also to the fields of transfer learning (TL) (see \cite{JMLR09-taylor} for a survey in the area of reinforcement learning). In this section we summarise some of the previous works in these fields which are related to our proposal. 

Structured Prediction (SP) is one example of learning from complex data context, where not only the input is complex but also the output. This has led to new and powerful techniques, such as Conditional Random Fields (CRFs) \cite{Lafferty:2001:CRF:645530.655813}, which use a log-linear probability function to model the conditional probability of an output $y$ given an input $x$ 
where Markov assumptions are used in order to make inference tractable. Other well-known approach is SVM for Interdependent and Structured Output spaces (SVM-ISO, also known as $SVM^{struct}$) seen as a SP evolution of \cite{Gaertner05} (Kernels) or \cite{vicentsim06,Tsochantaridis:2004:SVM:1015330.1015341,coin2012}. Also, \emph{hierarchical classification} can be viewed as a case of SP where taxonomies and hierarchies are associated with the output \cite{Koller:1997:HCD:645526.657130}.

Some of these previous approaches use special functions (probabilistic distributions, metrics or kernels) explicitly defined on the examples space. These methods either lack a model (they are instance-based methods) or the model is defined in terms of the transformed space. All these approaches lead to incomprehensible patterns/models in terms of the transformed (hyper-)space. A recent proposal which has tried to re-integrate the distance-based approach with the pattern-based approach is \cite{coin2012}, (leading, e.g., to Newton trees \cite{DBLP:conf/ausai/Martinez-PlumedEFHR10}).

Regarding the comprehensibility of patterns and their expressiveness power, inductive programming \cite{Kitzelmann10}, inductive logic programming (ILP) \cite{Mug99} and some of the related areas such as relational data mining \cite{DL01} are arguably the oldest and more successful attempts to handle complex data. They can be considered {\em general} machine learning systems, because any problem can be represented, preserving its structure, with the use of the Turing-complete languages underneath: logic, functional or logic-functional. Apart from their expressiveness, the advantage of these approaches is the capability of capturing complex problems in a comprehensible way. ILP, for instance, has been found especially appropriate for scientific theory formation tasks where the data are structured, the model may be complex, and the comprehensibility of the generated knowledge is essential. Learning systems using higher-order (see, e.g., \cite{Lloyd01knowledgerepresentation}) were one of the first approaches to deal with complex structures, which were usually flattened in ILP.

All these systems are based on the choice of fixed operators which leads to fixed heuristics. For instance, Plotkin's lgg \cite{Plo70} operator works well with a specific-to-general search. The ILP system Progol \cite{mug95} combines the Inverse Entailment with general-to-specific search through a refinement graph.  The Aleph system \cite{aleph} is based on Mode Direct Inverse Entailment (MDIE). In inductive functional logic programming, the FLIP system \cite{Ferri-RamirezHR01} includes two different operators: inverse narrowing \cite{hernandez1998inverse} and a consistent restricted generalisation (CRG) generator \cite{hernandez1999strong}. In any case, the set of operators configures and delimits the performance of each learning system. Also, rules that are learned on a first stage can be reused as background knowledge for subsequent stages (incremental learning). Hybrid approaches that combine genetic algorithms and ILP have also been introduced as in \cite{TM02}.

As an evolution of ILP into the fields of (statistical) (multi-)relational learning or related approaches, many systems have been developed to work with rich data representations. In \cite{springerlink:10.1007/s10994-008-5079-1}, for example, we can find an extensive description of the current and emerging trends in the so-called `structured machine learning' where the authors propose to go beyond supervised learning and inference, and consider decision-theoretic planning and reinforcement learning in relational and  the first-order settings.

There have been several approaches applying planning and reinforcement learning to structured machine learning  \cite{Tadepalli04}. While the term Relational Reinforcement Learning (RRL)  \cite{1007694015589,Tadepalli04} seems to come to mind, it offers state-space representation that is much richer than that used in classical (or propositional) methods, but its goal is not structured data. Other related approaches are, for instance, incremental models \cite{daume09searn,Maes2009SPRL} which try to solve the combinatorial nature of the very large input/output structured spaces since the structured output is built incrementally. These methods can be applied to a wide variety of techniques such as parsing, machine translation, sequence labelling and tree mapping.

A more general (and classical) approach, somewhat in between genetic algorithms and reinforcement learning, known as \emph{Learning Classifier Systems} (\emph{LCSs}) \cite{Holmes200223}. LCSs employ two biological metaphors: evolution and learning which are respectively embodied by the genetic algorithm, and a reinforcement learning-like mechanism appropriate for the given problem. Both mechanisms rely on what is referred to as the \emph{environment} of the system (the source of input data). In some way, the architecture of our system will resemble the LCS approach.

Learning to learn is one of the (desired) features of more general and flexible systems. One related area is meta-learning \cite{springerlink:Metalearning}.  Learning at the metalevel is concerned with accumulating experience on the performance of multiple applications of a learning system. A more integrated approach resembling meta-learning and incremental learning is  \cite{Schmidhuber:2004:OOP:969909.969942}, where the authors present the \emph{Optimal Ordered Problem Solver} (OOPS), an optimally fast way of incrementally solving each task in the sequence by reusing successful code from previous tasks, at least at a theoretical level.

Transfer learning (TL) is an area where experience gained in learning to perform one task can help improve learning performance in a related, but different, task. As the task and the learning system become more elaborated, this knowledge reuse becomes more important. There are three main families of TL methods (in the RL area) according to the description of states and actions.

Firstly, the source and target tasks use the same state variables and set of actions. The main idea is to break down a task into a series of smaller tasks. This approach can be considered a type of transfer in that a single large target task can be treated as a series of simpler source tasks and where transfer learning is performed by initialising the Q-values of a new episode with previously learned Q-values \cite{Carroll02fixedvs}. This family includes those TL methods which use multiple source tasks by leveraging all experienced source tasks when learning a novel target task \cite{Mehta05transferin} or by choosing a subset of previously experienced tasks \cite{Fernandez06probabilisticpolicy} 

The second family of methods are those which are able to transfer between tasks with different state variables and actions, 
 so that no explicit mapping between the tasks is needed. One approach is to make the agent reason over \emph{abstractions} of the Markov Decision Process  that are invariant when the actions or state variables change. Some methods use macro-actions or \emph{options} \cite{Sutton99betweenmdps}  to learn new action policies in Semi-Markov Decision Processes. This may allow the agent to leverage the sequence of actions needed to learn its task with less data. 
 \emph{Relational Reinforcement Learning} \cite{KDriessensThesis} is another particularly attractive formulation in this context of transfer learning. In RRL, agents can typically act in tasks with additional objects without reformulating them, although additional training may be needed to achieve optimal (or even acceptable) performance levels.

The last family 
 is more flexible than those previously discussed as they allow the state variables and available actions to differ between source and target tasks using inter-task mappings. Namely, explicit mappings are needed in order to transfer between tasks with different actions and state representations \cite{AAAI06-yaxin}. This inter-task mapping may be provided to the learner (\emph{advise} rules \cite{Price03acceleratingreinforcement}) or may be autonomously learned (\emph{qualitative dynamic Bayes networks} \cite{AAAI06-yaxin}). 


Given all the previous approaches to make machine learning methods more flexible in terms of data representation and reuse of previous learning experience, we consider the new idea of using operators and heuristics over them as the backbone of a more general machine learning paradigm.

\section{The \gerl system}\label{system}

\begin{figure}

	\centering
		\includegraphics[width=1.0\textwidth]{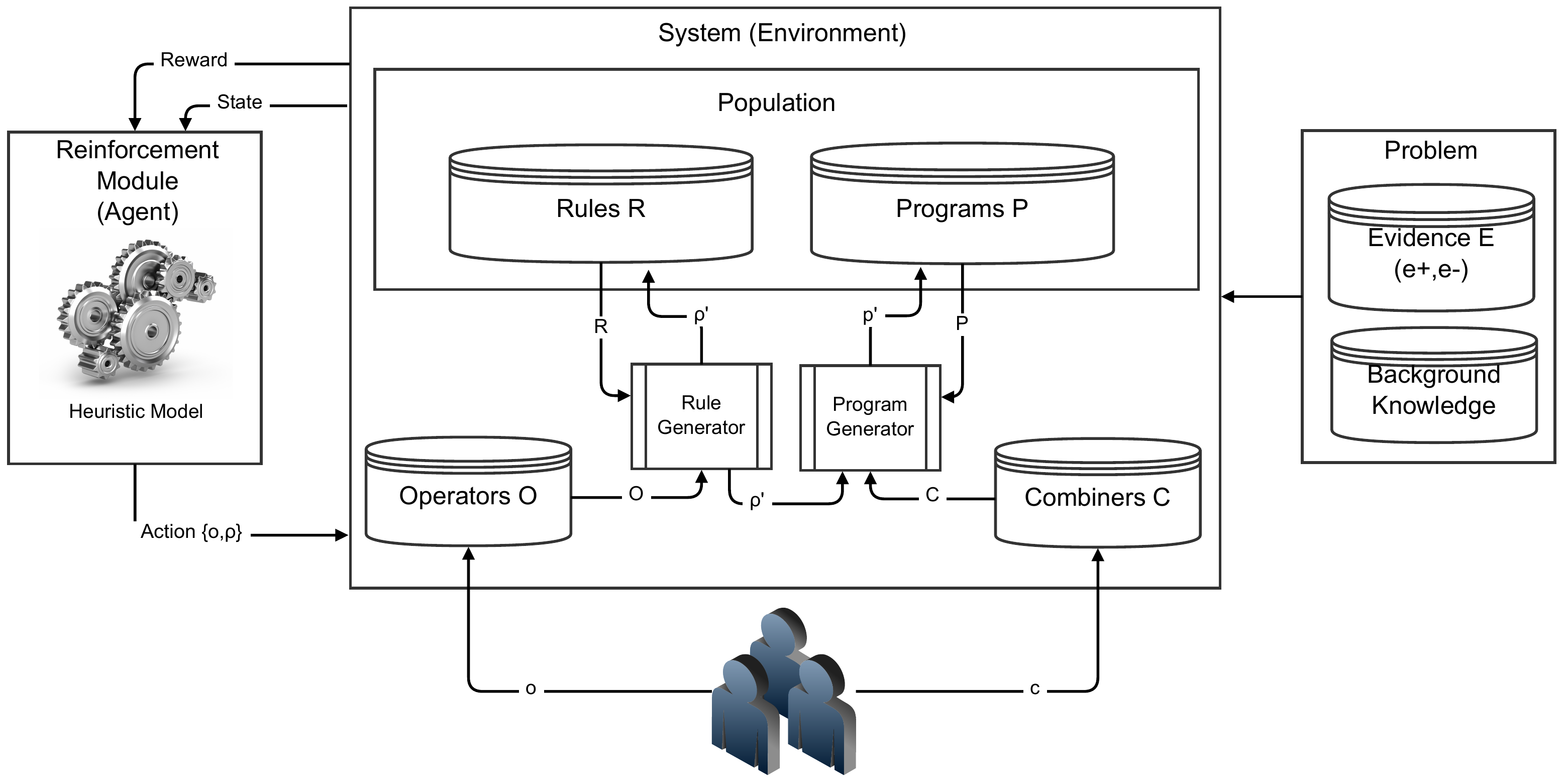}   
		
	\caption{{\gerl}'s system architecture}
	\label{fig:SystemArchitecture}

\end{figure}

As we have mentioned in Section \ref{intro}, in this paper we describe the \gerl system \cite{FmartinezNFMCP12} a general learning system which can be configured with different (possibly user-defined) operators and where the heuristics are also learnt from previous learning processes of (similar or different) problems. The system can be described as a flexible architecture (shown in Figure \ref{fig:SystemArchitecture}) which works with populations of rules (expressed as unconditional / con\-di\-tio\-nal equations) and programs in the functional language Erlang, which {\em evolve} as in an evolutionary programming setting or a learning classifier system \cite{Holland00}. Operators are applied to rules for generating new rules, which are then combined with existing or new programs. With appropriate operators, using some optimality criteria (based on coverage and simplicity) and using a reinforcement learning-based heuristic (where the application of an operator over a rule is seen as a decision problem fed by the optimality criteria) many complex problems can be solved.  As a result, this architecture can be seen as a `meta-learning system', that is, as a `system for writing machine learning systems' or to explore new learning operators.
In the rest of this section we will introduce some notation and concepts of the system that will be required to understand how our system has been devised and implemented.

\subsection{Why Erlang?}

Erlang/OTP \cite{Armstrong:1996:CPE:229883} is a functional programming language (which includes environment and libraries) developed by Ericsson (and used in production systems for over 20 years) and ``was designed from the ground up for writing scalable, fault-tolerant, distributed, non-stop and soft-realtime applications (like telecommunications systems)''. The core Erlang language consists of function definitions extended with message passing to handle concurrency, and OTP is a set of design principles and libraries that supports building fault-tolerant systems. 

Erlang, as a functional language, runs a program as a successive application of functions over an initial expression (free of variables). Branches of execution are selected based on pattern matching and loops are constructed using recursive functions. Once a variable is binding to a value, it cannot be changed. Most constructs (\emph{pure functions}) are side effect free, namely, functions that return the same value given the same arguments regardless of the context of the call of the function, exceptions are message passing and built-in functions (\emph{BIFs}). 
One interesting feature of Erlang is its strong dynamic nature. Variables are dynamically typed, there is no type checking at the compile-time. Function identifiers are a special data type called \emph{atom} and they can be generated at run-time and passed around in variables (higher order abilities). Execution threads are also created at run time, and they are identified by a dynamic system.

The main reasons why we have chosen \emph{Erlang} as the programming language of our system are: firstly, it is a free and open-source language with a large community of developers behind that implies that there is a large repository of libraries to deal, among other things, with the meta-level of the source code in an easy way (see $smerl$\footnote{ErlyWeb is a web development framework for Erlang:  {\ttfamily{https://code.google.com/p/erlyweb/}}} library); secondly, reflection and higher order, that allows us to interact easily with the meta-level representation of how rules and programs are transformed by operators; and finally, it is a unique representation language, which is appropriate for our requirements:  operators, examples, models and background knowledge are represented in the same language. The advantages of using the same representation language has been previously shown by the fields of ILP, IFP and IFLP (except for operators). 

Hence, we look for a flexible language, with powerful features for defining operators and able to represent all other elements (theories and examples) in an understandable way.

\subsection{\gerl Notation}

Let us introduce in this section the notation used for representing data and rules.

Let $\Sigma$ be a set of \emph{function symbols} together with their arity and $\cal{X}$ a countably set of \emph{variables}, then ${\cal{T}} (\Sigma,\cal{X})$ denotes the set of \emph{terms} built from $\Sigma$ and $\cal{X}$. Depending on the arity of symbols in $\Sigma$, a function  is said to be a {\em constant} if its arity=0 and otherwise it is said to be a {\em functor}. The set of variables occurring in a term $t$ is denoted \emph{Var(t)}. A term $t$ is a \emph{ground term} if $Var(t) = \varnothing$.  An equation is an expression of the form $l = r 
$ where $l$ (the left hand side, \emph{lhs}) and $r$ (the right hand side, \emph{rhs}) are terms. 
 ${\cal{R}}$ denotes the space of all (conditional) functional rules $\rho$ of the way $l\; [\hbox{when } G ] \rightarrow  B , r $
where $l$ and $r$ are the lhs and the rhs of $\rho$ (respectively),  $G=\{g_1,g_2, \dots g_m\;|\;m\geq0\})$ is a set of conditions or Boolean expressions called guards, and $B=b_1, \ldots, b_n$, the body of  $\rho$, is a sequence of equations. If $G = \varnothing $, then $\rho$ is said to be an unconditional rule. Let  ${\cal{P}}=2^{\cal{R}}$ be the space of all possible functional programs formed by sets of rules $\rho\in {\cal{R}}$. Given a program $p \in \cal{P}$, we say that term $t$ reduces to term $s$ with respect to $p$, $t \rightarrow_p s$, if there exists a rule $l\; [\hbox{when } G ] \rightarrow  B , r \in p$ such that a subterm of $t$ at occurrence $u$ matches $l$ with substitution $\theta$, all conditions $g_i\theta$ hold, for each equation $b_{i_l}=b_{i_r}\in B$, $b_{i_l}\theta$ and $b_{i_r}\theta$ have the same normal form (that is, $b_{i_l}\theta\rightarrow_p^* b$,  and $b_{i_r}\theta\rightarrow_p^* b$ and $b$ can not be further reduced) and $s$ is obtained by replacing  in $t$ the subterm at occurrence $u$ by $r\theta$. Sometimes, we will refer to  $B ,  r $ as $Right$.

An example $e$ is a ground rule $l\rightarrow r$ (that is, without condition nor body)  being $r$ in normal form  and both $l$ and $r$ are ground.   We say that $e$ 
is covered by a program $p$ (denoted by $p \models e$) if  $l$ and $r$ have the same normal form with respect to $p$, i.e. $l \rightarrow_p^* r$. A  program $p\in \cal{P}$ is a solution of a learning problem defined by a set of positive examples $E^+$, a (possibly empty) set of negative examples  $E^-$ and a background theory $K$ if it covers all positive examples, $K \cup p \models E^+$ (posterior sufficiency or completeness), and does not cover any negative example, $K \cup p \not\models E^-$ (posterior satisfiability or consistency). Our system has the aim of obtaining complete solutions, but their consistency is not a mandatory property, so approximate solutions are allowed. As usual, the coverage relation can also be  defined in terms of the operational mechanism of the functional language. We distinguish between positive and negative examples. Thus, the function $Cov^+: 2^{\cal{R}} \rightarrow {\mathbb{N}}$ calculates the positive coverage of a program $p\in 2^{\cal{R}}$ and it is defined as $Cov^+(p)=Card(\{e\in E^+: p \cup K \cup E^+ - e \models e  \})$, where $Card(S)$ denotes the cardinality of the set $S$. Analogously, the function $Cov^-: 2^{\cal{R}} \rightarrow {\mathbb{N}}$ calculates the negative coverage of a program $p\in 2^{\cal{R}}$ and it is defined as $Cov^-(p)=Card(\{e\in E^-: p \cup K \cup E^+ - e \models e\})$.  When we deal with recursive programs, we have to consider the problem of non-terminating proofs (that is, infinite sequences of rewriting steps). Note that, for proving the coverage of an example $e$ we use the set $(E^+ - \{e\})$ as base cases for the target function (this is known as extensional coverage). Moreover,  the length of the proofs are limited to a maximum number of rewriting steps.  



\begin{figure}

	\centering
		\includegraphics[width=0.6\textwidth]{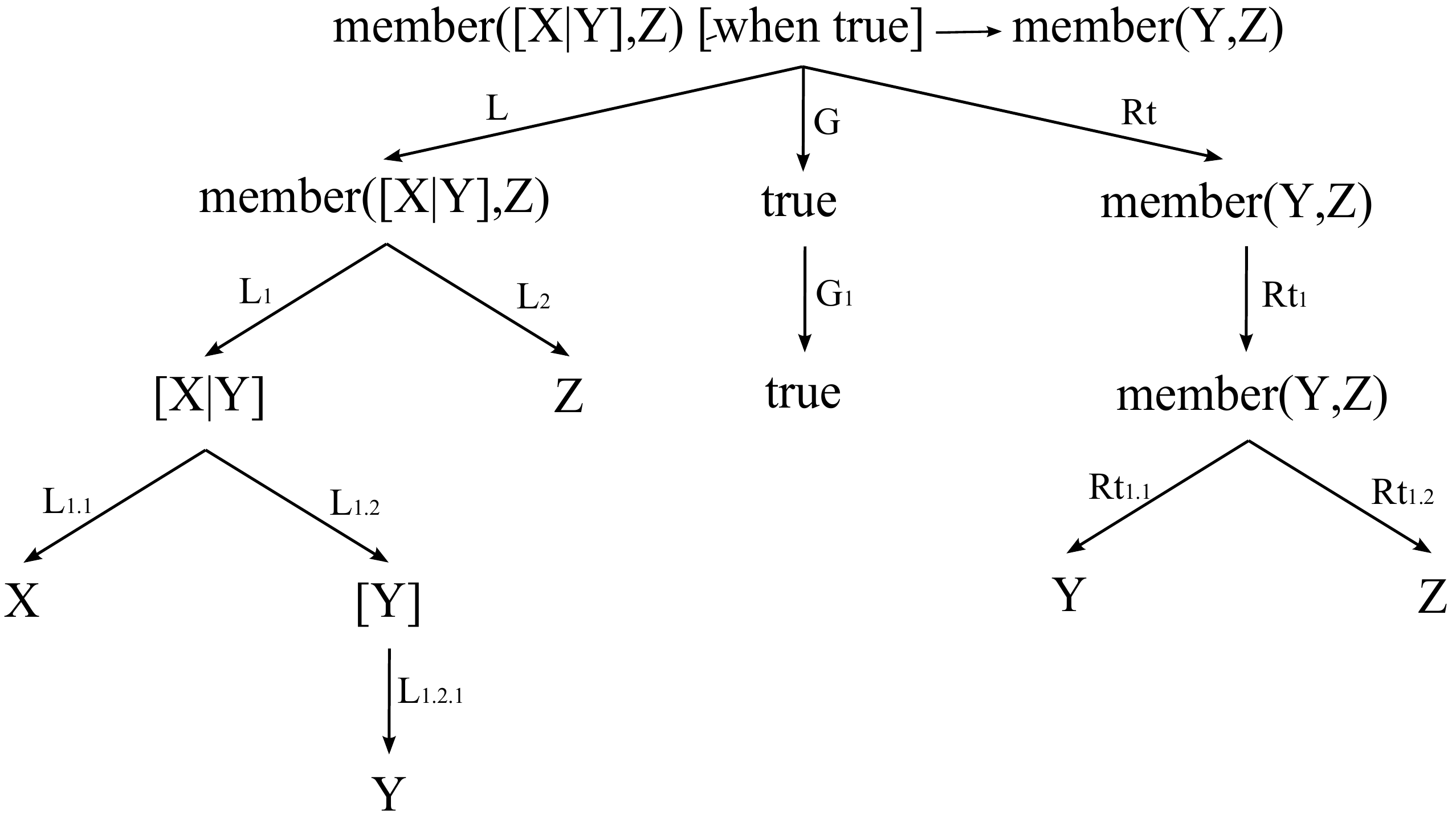}   
		
	\caption{Position tree of $member([X|Y],Z)\; when\;true \rightarrow member(Y,Z)$}
	\label{fig:PosTree}

\end{figure}

A rule $\rho$ can be represented as a tree, in a similar way as the usual tree representation of terms \cite{}. Given a rule $l\; [\hbox{when } G ] \rightarrow Right$, the root of the tree represents the complete rule, and its three children represent $l$, $G$ and $Right$.  Since $G$ and $Right$ are a conjunction of  guards and a conjunction of equations and one term, respectively, both nodes in the tree have as many children as components there are in the conjunctions. From here, the tree is populated following the usual tree representation of terms and equations. We call this kind of tree representation as {\em position trees}.  The position tree of a rule $\rho$ is used for denoting its subparts. In order to do it, the branches of the tree are labelled as follows: labels  $L$, $G$ and $Rt$ denote the nodes at depth $1$, that is,  the $lhs$, the guards and the  $(B,r)$ parts of 
$\rho$, respectively. Then, the labels of the nodes from depth 2  are obtained  by adding as subindexes of labels $L$, $G$ and $Rt$  a sequence of natural numbers (starting by 1) following a depth-first exploration of each subtree. Labels will be named positions. 
 Figure \ref{fig:PosTree} shows the position tree of the rule $\rho$: $member([X|Y],Z)\; when\;true \rightarrow member(Y,Z)$. As we can see, the subpart of $\rho$ at position $L_1$ is the term $[X|Y]$ and the subpart at position $Rt_{1,2}$ is the term $Z$. Abusing notation, we will use the position of a node for naming it.

It should be noted that the position tree of a rule $\rho$ matches exactly with the Erlang abstract representation of the rule, namely, the \emph{Abstract Syntax Tree} (AST)\footnote{Abstract Erlang syntax trees description: {\ttfamily{http://erlang.org/doc/man/erl\_syntax.html}}} which is a tree representation of the abstract syntactic structure of a source rule written in Erlang. Each node of the tree denotes a component occurring in the source code. If we parse the abstract tree, we can enumerate each component in a post-order way obtaining a position tree.

\comment{
\begin{figure}
	\centering
		\includegraphics[width=0.7\textwidth]{Positions.pdf}
	\caption{Positions of a rule $\rho$}
	\label{fig:pos}
\end{figure}
}

Hereinafter, we will consider the following running example in order to illustrate how the system works. This example will be developed along the following sections.

\begin{exmp}\label{ex}
Consider a simple recursive problem of finding the last element of a list of characters. This problem could be defined using a set of positive examples $E^+$ and a set of negative ones $E-$ (and an empty set of functions as the background knowledge $K$) as in Table \ref{tab:lastEx}.

\begin{table}

\centering
{\sffamily\small
\begin{tabular}{clcl}
\hline 
\textbf{id} & $\mathbf{E^+}$	& \textbf{id}		& $\mathbf{E^-}$ 								\\
\hline

1 & $last([c])\rightarrow c.$ 				& 1 & $last([c])\rightarrow b.$					\\
2 & $last([d])\rightarrow d.$ 				& 2 & $last([b])\rightarrow l.$					\\
3 & $last([l])\rightarrow l.$ 				& 3 & $last([l])\rightarrow c.$				\\
4 & $last([a,b,c])\rightarrow c.$ 		& 4 & $last([a,b,c])\rightarrow a.$			\\
5 & $last([t,b,n,a,b])\rightarrow b.$ & 5 & $last([t,b,n,a,b])\rightarrow t.$	\\
6 & $last([h,h,t,a,l])\rightarrow l.$ &   & 																	\\
7 & $last([a,c,b])\rightarrow b.$ 		&	  & 																	\\
8 & $last([a,b,a,c])\rightarrow c.$ 	&	  & 																	\\

\hline

\end{tabular}
}

\caption{Set of positive and negative examples provided to \gerl in order to learn the recursive problem of \emph{last element of a list}.}
\label{tab:lastEx}

\end{table}

\end{exmp}

\subsubsection{Operators over rules and programs}\label{sec:operators}

The definition of customised operators is one of the key concepts of our proposal. In \gerl, rules $R$ are transformed by applying a set  of \emph{operators} $O $. Then, an operator $o\in \cal{O}$ is  a function $o: {\cal{R}} \rightarrow 2^{\cal{R}}$,  where $O \subseteq \cal{O}$ denote the set of operators chosen by the user for solving the problem.
Roughly speaking, the operator's aim is to perform modifications over any of subparts of a rule in order to generalise or specialise it. The main idea is that, when the user is going to deal with a new problem, he/she can define his/her own set of \emph{operators} (which can be selected from the set of predefined operators or can be defined by the user with the functions provided by the system) especially suited for the data structures of the problem. This feature allows our system to adapt to the problem at hand.

For defining operators, the system is equipped  with meta-level facilities called {\em meta-operators}. A meta-operator is formally defined as \[\mu O :: Pos \times {\cal{T}}(\Sigma,\cal{X}) \rightarrow {\cal{O}}\] 

\noindent which takes a position in a rule (as given by its position tree) and a term, and gives an operator. \gerl provides the following three meta-operators able to define well-known generalisation and specialisation operators in Machine Learning:

\begin{enumerate}
	
	\item $ meta\_replace (Pos,  Term) $: creates an operator that replaces in a rule a term located in the position $Pos$ by the term $Term$. Notice that this meta-operator can be used to define both generalisation (replacing, for instance, a term by a variable) and specialisation operators (replacing a given term by another more specific).
	
	\item $ meta\_insert (Pos, Term) $: creates an operator that inserts a term $Term$ in the position $Pos$ of a rule. Notice that this meta-operator can be only used to define specialisation operators (we specialise a rule adding more terms to its body).
	
	\item $ meta\_delete (Pos)$: creates an operator that deletes a term located in the position $Pos$ of a rule. Notice that this meta-operator can be only used to define generalisation operators (deleting guards from $G$ or equations from $Right$).
	
\end{enumerate}


\comment{
Let us see an example. Given a rule $FName(Arguments)$ $\rightarrow RHS$,  where $Arguments$ is a list,  imagine that we want to define an operator for obtaining the head of $Arguments$ and return it as the rhs of a new rule. This operator could be defined as:

$$replace(Fname(Arguments) \rightarrow RHS, R_1, head(L_1))$$ 
$$\Rightarrow Fname(Arguments) \rightarrow head(Arguments)$$

\noindent where $\Rightarrow$ represents the rule transformation relation defined by the operators and $head$ is a BiF (\emph{Built-in Function}) of Erlang, and take $Arguments$ (which is a list located in position $L_1$) as input.

}

\begin{figure}[ht]
\begin{minipage}[b]{0.49\linewidth}
\centering
\includegraphics[width=\textwidth]{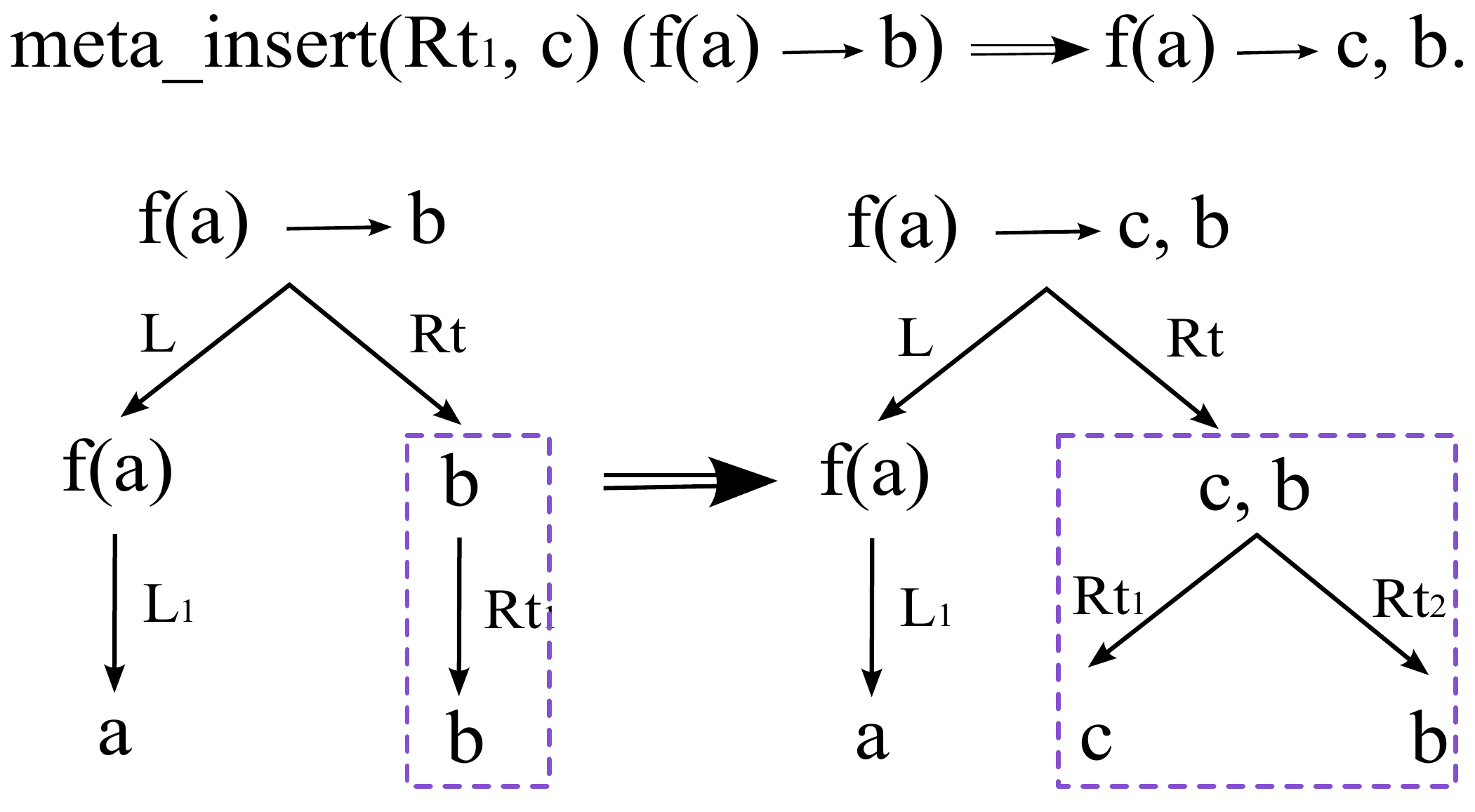}
\end{minipage}
\hspace{0.5cm}
\begin{minipage}[b]{0.48\linewidth}
\centering
\includegraphics[width=\textwidth]{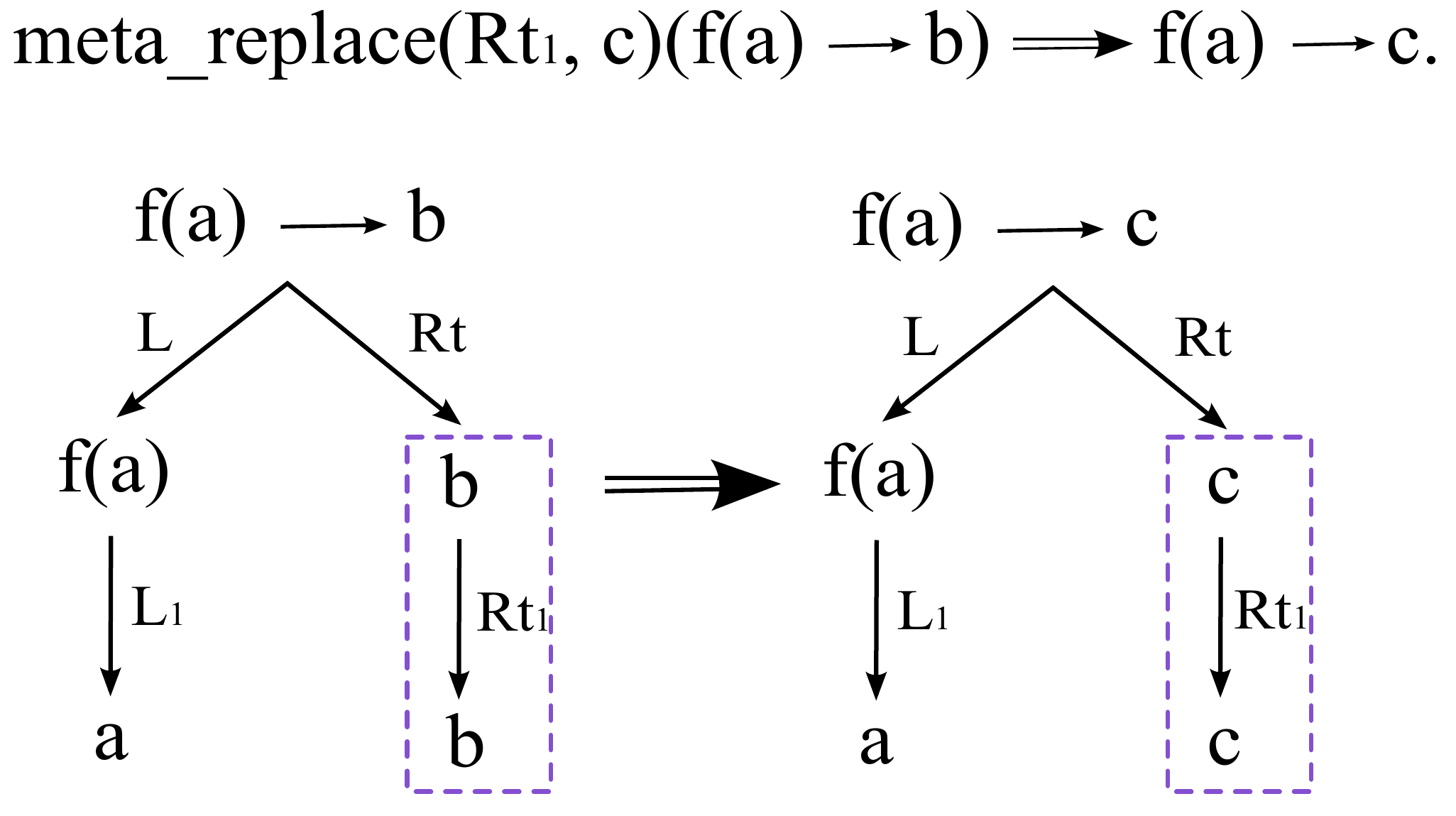}
\end{minipage}
\caption{Difference between the meta-operators $meta\_insert$ and $meta\_replace$. (Left) Example of transformation of the Position Tree after applying the meta-operator $meta\_insert$. (Right) Example of transformation of the Position Tree after applying the meta-operator $meta\_replace$.}\label{fig:difmet}
\end{figure}

\noindent Both meta-operators can also be used in order to generate recursive rules by  just providing a second argument of the meta-operator being a term with the target function as the outermost symbol. On the other hand, we would like to highlight that the meta-operator $meta\_insert$ has not the same meaning than $meta\_replace$, although both of them can be used to define specialisation operators. The first one moves the elements which are on the right of $Pos$ one position to the right  and inserts a new term in $Pos$, while $meta\_replace$ substitutes the term in $Pos$ directly.  In Figure \ref{fig:difmet} we illustrate this difference with an example in that we apply the meta-operators for inserting  and replacing the term $c$ in the position $Rt_1$ of the rule $\rho = f(a) \rightarrow b$.

Finally, there is an  internal operator called $one\_step\_rew$ which, by default,  is always included  in  the set $\cal{O}$. This operator performs a step of rewriting at any position of the $Right$ component of a rule.

\comment{
$replace({\textcolor{orange}{FName}}({\textcolor{red}{Arguments}})\rightarrow {\textcolor{blue}{RHS}})\ {\textcolor{violet}{[when\  Arguments\  is\ a\  List]}} \\
\hbox{~~~~~~~~~~~}\Rightarrow ({\textcolor{orange}{FName}}({\textcolor{red}{Arguments}})\rightarrow {\textcolor{green}{head({\textcolor{red}{Arguments}})}}).$\\\noindent 

\noindent where $\Rightarrow$ represents the rule transformation relation defined by the operators. The codification in Erlang could be as follows:

{\ttfamily{
 \noindent  Operator\_takeHead({\textcolor{BlueGreen}{Rule}}) -> \\
(1) \indent  \{function,\_,{\textcolor{orange}{FName}},\_,\{clause,\_,{\textcolor{red}{Arguments}},{\textcolor{violet}{Guards}},{\textcolor{blue}{RHS}}\}\} = {\textcolor{BlueGreen}{Rule}},\\
(2) \indent  \{cons,\_,{\textcolor{green}{L1}},L2\} = {\textcolor{red}{Arguments}},\\
(3) \indent	 \{function,\_,{\textcolor{orange}{FName}},\_,\{clause,\_,{\textcolor{red}{Arguments}},{\textcolor{violet}{Guards}},{\textcolor{green}{L1}}\}\}.\\
}}
%

\noindent where identifiers with a capital letter followed by any combination of uppercase and lowercase letters and underscores are Erlang variables, and other static (or constants) literals are Erlang atoms. In line 1,  the {\ttfamily{\textcolor{BlueGreen}{Rule}}} is parsed and transformed into a valid Erlang  \emph{abstract syntax tree} (AST)  in order to easily access to its components: the Erlang \emph{forms} {\ttfamily{{\textcolor{orange}{FName}}, \textcolor{red}{Arguments}}, {\textcolor{violet}{Guards}} and {\textcolor{blue}{RHS}}}. Next, the operator decomposes  {\ttfamily{\textcolor{red}{Arguments}}} into the Erlang meta-expression for lists (line 2), and finally, line 3 returns the new AST constructed by replacing   the {\ttfamily{\textcolor{blue}{RHS}}} part by {\ttfamily{\textcolor{green}{L1}}} in  the AST obtained in line 1. For simplicity, we have omitted some further code for checking the arity and type of {\ttfamily{\textcolor{red}{Arguments}}}.  
}

Our system also has a special kind of transformation $c \in C$, called \emph{combiners}, that only apply to programs. The \emph{Program Generator} module (Figure \ref{fig:SystemArchitecture}) applies a combiner to the last rule $\rho'$ generated by the {\em Rule Generator}  module and the population of programs $P$. Thus, a combiner $c \in C$  can be formally described as a function $c: {\cal{P}} \times {\cal{P}} \rightarrow {\cal{P}}$ that transforms programs into programs.

Following the previous running example \emph{last element of a list} defined in Example \ref{ex} as a set of positive and negative evidence, the next step is to define appropriate operators (relying on the previous meta-operators) in order to allow the system to learn possible solutions. Knowing that a list could be navigated in a recursive way, it is easy to see that we need to insert recursive calls into the $Right$ part of the rules taking as input the head or the tail of the input list, namely:

\[\begin{array}{lll} 

\blue[op_1]\equiv & meta\_replace(Rt_1,last(L_{1.1})) 	& \text{\scriptsize \ttfamily {// Recursive call takes the}}\\ 
									&																			& \text{\scriptsize \ttfamily {// head of the input list}}\\ 
\green[Example]: & \blue[op_1](last([a,b,a,c])\rightarrow c)  \Rightarrow & last([a,b,a,c])\rightarrow last(a) \\

\blue[op_2]\equiv & meta\_replace(Rt_1,last(L_{1.2})) 	& \text{\scriptsize \ttfamily {// Recursive call takes the }}\\
									& 																		&	 \text{\scriptsize \ttfamily {// tail of the input list}} \\
\green[Example]: & \blue[op_2](last([a,b,a,c])\rightarrow c)  \Rightarrow & last([a,b,a,c])\rightarrow  last([b,a,c]) \\

\end{array}\]

Seen the evidence, it can be useful to define some operators which can play with the structure of the list replacing the $Right$ part of the rules with each part of the lists:

\[\begin{array}{lll} 

\blue[op_3]\equiv & meta\_replace(Rt_1,L_{1.1}) 	& \text{\scriptsize \ttfamily {// Takes the head of the input list}}\\ 
\green[Example]: & \blue[op_3](last([a,b,a,c])\rightarrow c) \Rightarrow & last([a,b,a,c])\rightarrow  a \\

\blue[op_4]\equiv & meta\_replace(Rt_1,L_{1.2}) 	& \text{\scriptsize \ttfamily {// Takes the tail of the input list}} \\
\green[Example]: & \blue[op_4](last([a,b,a,c])\rightarrow c)  \Rightarrow & last([a,b,a,c])\rightarrow  [b,a,c] \\

\end{array}\]

Finally, in order to generate general rules, we need to define operators that replace both the head and the tail of the input lists by a variable:

\[\begin{array}{lll} 

\blue[op_5]\equiv & meta\_replace(L_{1.1},V_{Head}) 	& \text{\scriptsize \ttfamily {// Generalise the head of the }}\\
									&																		&	\text{\scriptsize \ttfamily {// input list}}\\ 
\green[Example]: & \blue[op_5](last([a,b,a,c])\rightarrow c.) \Rightarrow & last([V_{Head},b,a,c])\rightarrow c \\

\blue[op_6]\equiv & meta\_replace(L_{1.2},V_{Tail}) 	& \text{\scriptsize \ttfamily {// Generalise the tail of the }} \\
									&																		&	\text{\scriptsize \ttfamily {// input list}}\\ 
\green[Example]: & \blue[op_6](last([a,b,a,c])\rightarrow c)  \Rightarrow & last([a,V_{Tail}])\rightarrow c \\

\end{array}\]

\section{RL-based heuristics}\label{heuristic}

In this section we describe the reinforcement learning approach followed by \gerl in order to guide the learning process. The freedom given to the user concerning the definition of their own operators implies the impossibility of defining specific heuristics to explore the search space. This means that heuristics must be overhauled as the problem of deciding the operator that must be used (over a rule) at each particular state of the learning process. 

A Reinforcement Learning (RL) \cite{sutton1998reinforcement} approach suits perfectly for our purposes, where the population of rules at each step of the learning process can be seen as the \emph{state} of the system and the selection of the tuple operator and rule can be seen as the \emph{action}. However, the probably infinite number of states and actions makes the application of classical RL algorithms not feasible. To overcome this, states and actions are represented in an abstract way using features. From here, a model-based Reinforcement Learning approach has been developed in order to use propositional machine learning methods for selecting the best action in each possible state of the system. 

\subsection{Optimality and stop criterion}\label{optimality}

Since the system is flexible and general in the way it represents and operates with rules, we need some general optimality criteria. First we have to select the optimal program (or a set of optimal programs, depending on the user's interests) as the solution of the learning problem. Second, we also need to feed the reward module in each step of the learning process.

The \emph{Minimum Message Length} \cite{Wallace01081968} (MML) provides a Bayesian interpretation of the Occam's Razor principle: the model generating the shortest overall  message (composed by the model and the evidence concisely encoded using it)  is more likely to be correct.  The MML principle is one of the most popular selection criterion in inductive inference (for a formal justification and its relation to Kolmogorov complexity and the related MDL principle, see  \cite{Li:2008:IKC:1478784, DBLP:journals/cj/WallaceD99a,Wallace:2005:SII:1051763}). 


According to the MML philosophy, the optimality of a program $p$ is defined as the weighted sum of two simpler heuristics, namely, a complexity-based heuristic (which measures the complexity of $p$) and a coverage heuristic (which measures how much extra information is necessary to express the evidence given the program $p$): 

\begin{equation}	 				 
Opt(p) = -\beta_1\cdot MsgLen(p)  - \beta_2\cdot MsgLen(\left\langle E^+,E^- \right\rangle | p )  \label{Opt} 
\end{equation}



\noindent Since programs are composed by rules, the coding of $p$  (in bits) can be approximated from the number of symbols that occur in each one of the rules. More concretely, if $\Sigma$ is composed by $n_f$ functors and $n_c$ constants and $n_v$ variable symbols, we define $MsgLen(p)$ as:

\begin{equation}
MsgLen(p)= \sum_{\rho \in p}{f*log_2(n_f+1) + c*log_2(n_c+1)+ v* log_2(n_v+1)}\label{msglen} 
\end{equation}

\noindent where $f$, $c$ and $v$ are the number of functors, constants and variables in rule $\rho$, respectively. 

The coding lengths of the positive instances not covered and the negative instances covered by $p$:

\begin{eqnarray}
MsgLen(\left\langle E^+,E^-  \right\rangle | p) = MsgLen(\{e \in E^+ : p \not\models e\}) \nonumber \\
+ MsgLen(\{e \in E^- : p \models e\}) \label{covf} 
\end{eqnarray}

As we have mentioned at the beginning of this section, the optimality measure is used to rank the programs generated by the system (in increasing order of their optimalities). 

\begin{figure}
	\centering
		\includegraphics[width=0.6\textwidth]{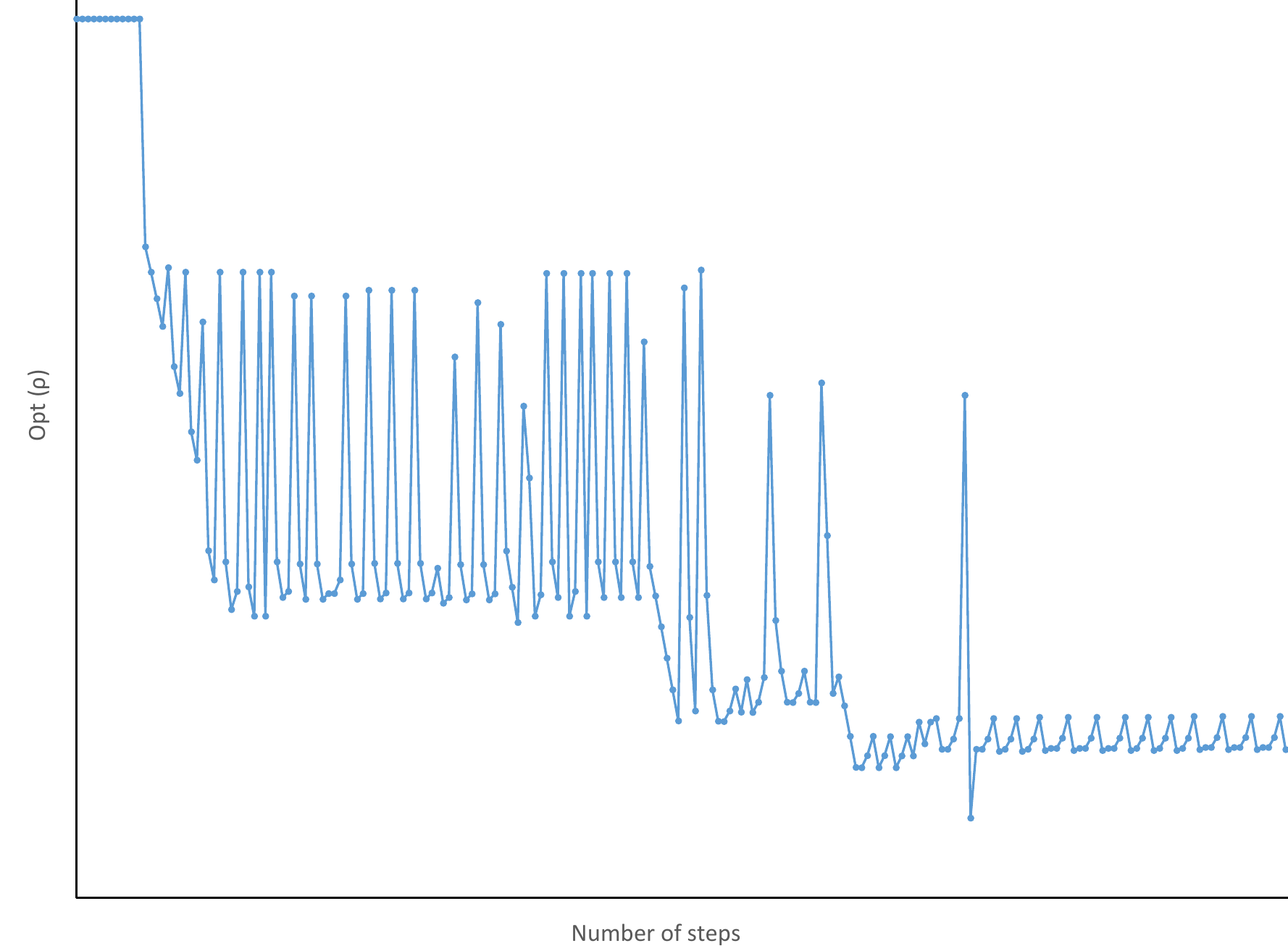}
	\caption{Example of the optimality values of the programs generated by \gerl in successive steps (starting from 1). Notice that in final steps, the variation of the optimality is not constant but shows a stabilisation.}
	\label{fig:stop}
\end{figure}

Finally, regarding the \emph{stop criterion}, it can be specified by the combination of two conditions. The first one establishes that the learning process stops  when the difference between the optimalities  of the programs generated in the last $n$ steps (where $n$ is a parameter determined by the user)  is less than a threshold  $\epsilon$ (also determined by the user) which indicates that a better program are not likely to be found.   Figure \ref{fig:stop}) shows the optimality of the programs generated at each step of the system for a given problem. As can be seen, the difference between the optimalities of the last generated programs fluctuates inside a tight margin, so the system stops by condition 1. The second stop condition limits the number of steps of learning process to a maximum.

\subsection{RL problem statement}\label{states}

As we can see in Figure \ref{fig:SystemArchitecture}, \gerl works with two sets: a set of rules $R\subseteq \cal{R}$  and a set of programs $P\subseteq \cal{P}$, where each program $p \in P$ is composed by rules belonging to $R$. Initially, $R$ is  populated with the positive evidence  $E^+$ and the set of programs $P$ is populated with as many unitary programs as rules are in $R$. As the process progresses, new rules and programs will be generated. First, the {\em Rule Generator} module  (Figure \ref{fig:SystemArchitecture}) takes the operator $o$ and the rule $\rho$ returned 
by the \emph{Reinforcement Learning Module} (policy). Then the operator is applied to the rule giving new rule $\rho'$ 
which are added to $R$. 
Then, the \emph{Program Generator} module takes the new generated rule $\rho'$ (if appropriate), the set of programs $P$ and the set of combiners $C$ and generates a new program $p'$ (which is added to $P$). Therefore, in each iteration of the system, we have to select a rule and an operator to produce new rules. Depending on the problem to solve, the number of required iterations to learn a problem could be astronomically high. To address this issue we need a particular heuristic (policy) to tame the search space and make good decisions about the choice of rule and operator, in which the application of an operator to a rule is seen as a decision problem. 

To guide the reinforcement learning process we need to describe the system in each step of the process (before and after applying an action) in terms of the quality of the system states (that is, the population of rules and programs generated until now).  
The infinite number of states makes their abstraction necessary. As there are infinitely many rules, we also have to use an abstraction for actions. This is done by defining an abstract representation of states and actions which constitutes the configuration of the RL problem we see next. %


Formally, we define a state at each iteration or step $t$ of the system as a tuple $\sigma_t=\langle R,P \rangle$ which represent the population of rules $R$ and programs $P$ in $t$.   An action is a tuple $\langle o, \rho \rangle$ with   $o \in {\cal{O}}$ and $\rho \in {\cal{R}}$ that represents the operator $o$ to be applied to the rule $\rho$. 
 Our decision problem  is a four-tuple $\langle {\cal{S}},{\cal{A}},\tau,\omega \rangle$ where: $\cal{S}$ is the state space; 
 $\cal{A}$ is a finite actions space ($\cal{A} = \cal{O} \times \cal{R}$); $\tau: {\cal{S}} \times {\cal{A}} \rightarrow {\cal{S}}$ is a transition function between states and $\omega : {\cal{S}} \times {\cal{A}} \rightarrow \mathbb{R}$  is the reward function. These components are defined below:

\begin{itemize}
	\item {\bf{States.}}	As we want to find a good solution to the learning problem, we represent each state $s_t$ by a tuple of features $\dot{s}_t=\langle \phi_1, \phi_2, \phi_3 \rangle$ from which to extract relevant information in step $t$:

\begin{enumerate}

	\item \emph{Global optimality} $(\phi_1)$: this  factor is   calculated as the average of the optimalities of all programs in the system at step $t$, denoted by $p_t$(using equation\ref{Opt}):

 				\begin{equation}
	 				 GlobalOpt(P_t) = \frac{1}{Card(P_t)}  \sum_{p \in P_t} Opt(p) 		\label{OptGlobal}	 
				 \end{equation}

	\item \emph{Average Size of Rules} $(\phi_2)$: measures the average complexity of all rules in $R$, using the measure function denoted in eq. \ref{msglen}. 
		
	\item \emph{Average Size of programs} $(\phi_3)$: measures the average cardinality of all the programs in $P_t$ in terms of the number of rules.

\end{enumerate}

\noindent We denote by $\dot{{\cal{S}}} = \mathbb{R}^3$ the set of state abstractions used to represent elements ${\cal{S}}$ in $\dot{{\cal{S}}}$. 
	
	\item {\bf{Actions.}} An action is a tuple $\langle o, \rho \rangle$ where $o$ is just an operator identifier (it is not abstracted) and each rule $\rho$ is described by a tuple of features $\dot{\rho}=\langle \varphi_1, \varphi_2, \varphi_3, \varphi_4, \varphi_5, \varphi_6, \varphi_7, \varphi_8 \rangle$ from which we extract relevant information:
	
			\begin{enumerate}
				\item \emph{Size ($\varphi_1$)}: expressiveness of the rule using \ref{msglen}. 
				
				\item  \emph{Positive Coverage Rate ($\varphi_2$)}. 
				
				\item  \emph{Negative Coverage Rate ($\varphi_3$)}. 
				
				\item  \emph{NumVars ($\varphi_4$)}: number of variables of $\rho$.
				
				\item  \emph{NumCons ($\varphi_5$)}: number of constants (functors with arity 0) of $\rho$.
				
				\item  \emph{NumFuncs ($\varphi_6$)}: number of functors with arity greater than 0 of $\rho$.
				
				\item  \emph{NumStructs ($\varphi_7$)}: number of structures (lists, graphs, \dots) of $\rho$.
				
				\item  \emph{isRec ($\varphi_8$)}: indicates if the rule $\rho$ is recursive or not.

			\end{enumerate}
	\noindent As an action consists of a choice of operator and rule, an action is finally abstracted as a tuple of {\em nine} features, i.e. $\dot{{\cal{A}}}=\left\langle \mathbb{N}, \mathbb{R}^8 \right\rangle$, where the abstraction of the actions goes from ${\cal{A}} \rightarrow \dot{{\cal{A}}}$. 
	
	\item {\bf{Transitions.}} Transitions are deterministic. A transition $\tau$ evolves the current sets of rules and programs by applying the operators selected (together with the rule) and the combiners.
	
	\item {\bf{Rewards.}} The optimality criterion seen above (eq. \ref{Opt}) is used to feed the rewards.
	
	\end{itemize}

\noindent At each point in time, the reinforcement learning policy $\pi$ can be in one of the states $s_t \in {\cal{S}}$ and may select an action  $a_t \in {\cal{A}}$ to execute. Executing such action $a_t$ in $s_t$ will change the state into $s_{t+1} = \tau(s_t,a_t)$, and the policy receives a reward $w_t=\omega(s_t,a_t)$. The policy does not know the effects of the actions, i.e. $\tau$ and $\omega$ are not known by the policy and need to be learned. This is the typical formulation of reinforcement learning \cite{sutton98a} but here we use features to represent the states and the actions. With all these elements, the aim of our decision process is to find a policy $\pi: {\cal{S}} \rightarrow {\cal{A}}$ that maximises:

\begin{equation}
		V^{\pi}(s_t)=\sum_{i=0}^{\infty} \gamma^i w_{t+i}
\end{equation}

\noindent for all $s_t$, where $\gamma \in [0,1]$ is the \emph{discount parameter} which determines the importance of the future rewards ($\gamma = 0$ only considers current rewards, while $\gamma = 1$ strives for a high long-term reward).


\subsection{Modelling the state-value function: using a regression model}


In our system, as we work with an abstract representation of states and actions,  we use a hybrid between  value-function methods (which update a state-value matrix)
and model-based methods (which learn models for $\tau$ and $\omega$) \cite{sutton98a}. 
The idea is to replace the \emph{state-value} function $Q(s,a)$ of the Q-learning \cite{springerlink:10.1007/BF00992698} (which returns quality \emph{values}, $q \in \mathbb{R}$) 
by a supervised model
$Q_M: \dot{{\cal{S}}} \times \dot{{\cal{A}}} \rightarrow \mathbb{R}$ that calculates the $q$ value for each state and action, using their abstractions. 
 \gerl uses linear regression (by default, but other regression techniques could be used) for generating $Q_M$, which is retrained  periodically from $Q(s,a)$. 
Then, $Q_M$ is used to obtain the best action $a_t$ for the state $s_t$ as follows: \\

\begin{equation}
a_t=\arg\max_{a \in {\cal{A}}}\left\{Q_M(\dot{s}_t,\dot{a}) \right\} \label{maxa} 
\end{equation}

Since the rules are described in an abstract way, more than one rule can share the same description, if that happens when the system selects an action, the rule is randomly selected (among those which share the description).

\begin{figure}[htbp]
	\centering
		\includegraphics[width=1.00\textwidth]{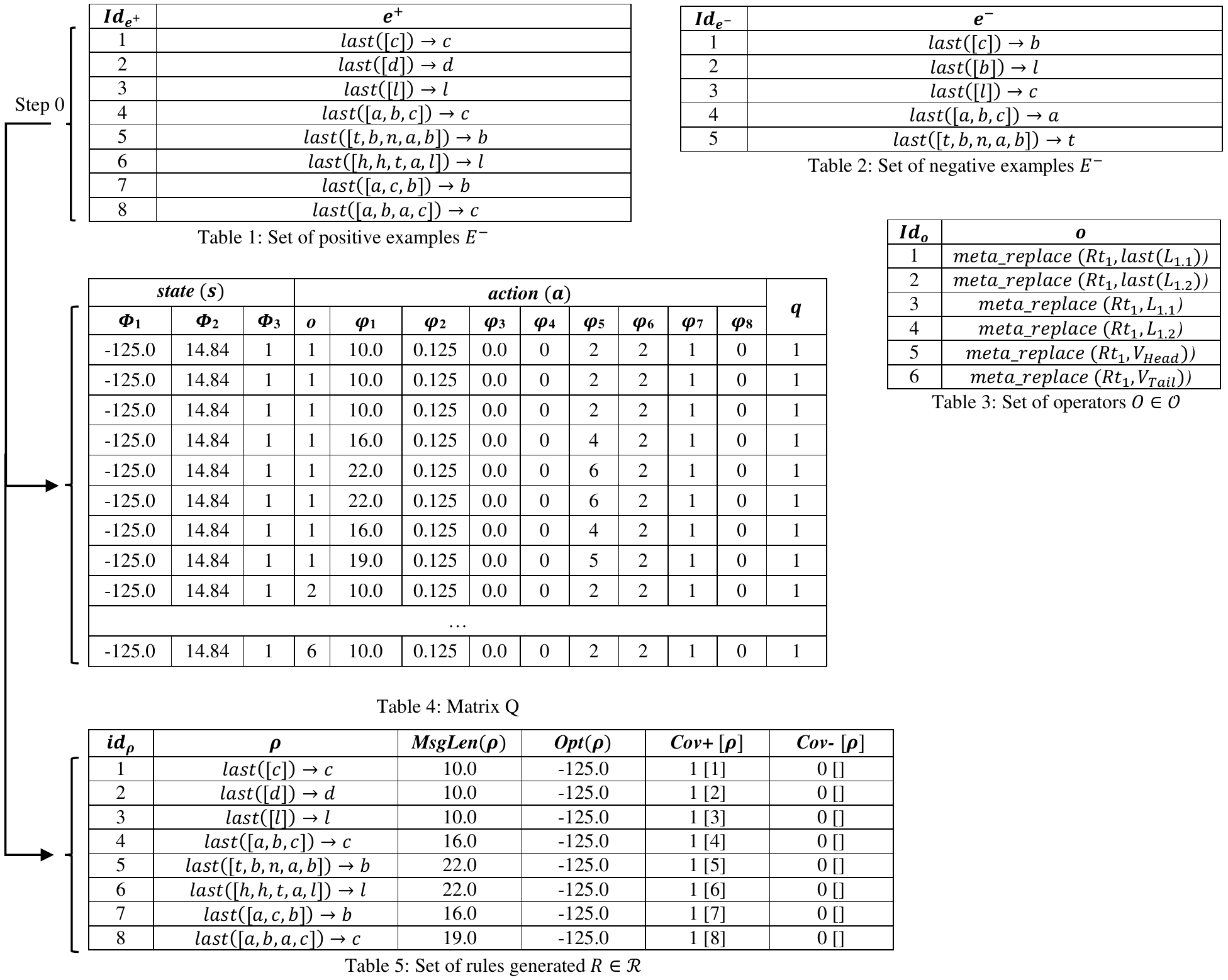}
	\caption{\gerl learning the \emph{last element of a list} problem: in step 0, R is initialised with the positive examples, and Q is initialised with all combination of actions abstractions (operators $\times$ rules) with $q$ values equal to 1. }
	\label{fig:problemInstanceVerticalA}
\end{figure}

In order to train the model we need to provide different states and actions as inputs, and quality values as outputs. More concretely, 
we use a `matrix' $Q$ (which is actually a table), whose rows are in $\dot{{\cal{S}}} \times \dot{{\cal{A}}} \times \mathbb{R}$ where $\dot{{\cal{S}}}$ is a tuple of state features as described in section \ref{states}, $\dot{{\cal{A}}}$ is the tuple of rule features and operator as also described in section \ref{states}, and $\mathbb{R}$ is a real value for $q$.
Abusing notation, to work with $Q$ as a function (like the original $Q$-matrix is used in many RL methods), we will denote by $Q[\dot{s},\dot{a}]$ the value of $q$ in the row of $Q$ for that state $\dot{s}$ and action $\dot{a}$. 
So, $Q$ grows in terms of the number of rows. Figure \ref{fig:problemInstanceVerticalA} continues with the example \ref{ex} and shows how \gerl initialises the $Q$ table and the set of rules $R$. 

 Once the system has started, at each step, $Q$ is updated using the following formula:

	\begin{equation}
			Q[\dot{s}_t,\dot{a}_t] \leftarrow \alpha \left[ w_{t+1} + \gamma \max_{a_{t+1}} Q_{M}(\dot{s}_{t+1},\dot{a}_{t+1}) \right] + (1-\alpha) Q[\dot{s}_t,\dot{a}_t] \label{upQ}
	\end{equation}
	
\noindent which is a variation of the formula used in Q-learning for updating the Q-matrix where the maximum future value is given by the model. 
The previous formula has two parameters: the discount parameter $\gamma \in [0,1]$, and the \emph{learning rate} $\alpha$ ($\alpha \in [0,1]$) which determines to what extent the newly acquired information will override the old information 	($\alpha = 0$ makes the agent not to propagate anything, while $\alpha = 1$ makes the agent consider only the most recent information). By default, $\alpha = 0.5$ and $\gamma = 0.5$.  

Following with example \ref{ex} , Figure \ref{fig:problemInstanceVerticalA} and ref{fig:problemInstanceVerticalB} show how \gerl uses $Q_M$ in order to get the best action to apply in each step of the learning process, and how the set of rules $R$ and the Q-matrix are updated until the system reaches the \emph{Stop Criterion}.

\begin{figure}[htbp]
	\centering
		\includegraphics[width=1.00\textwidth]{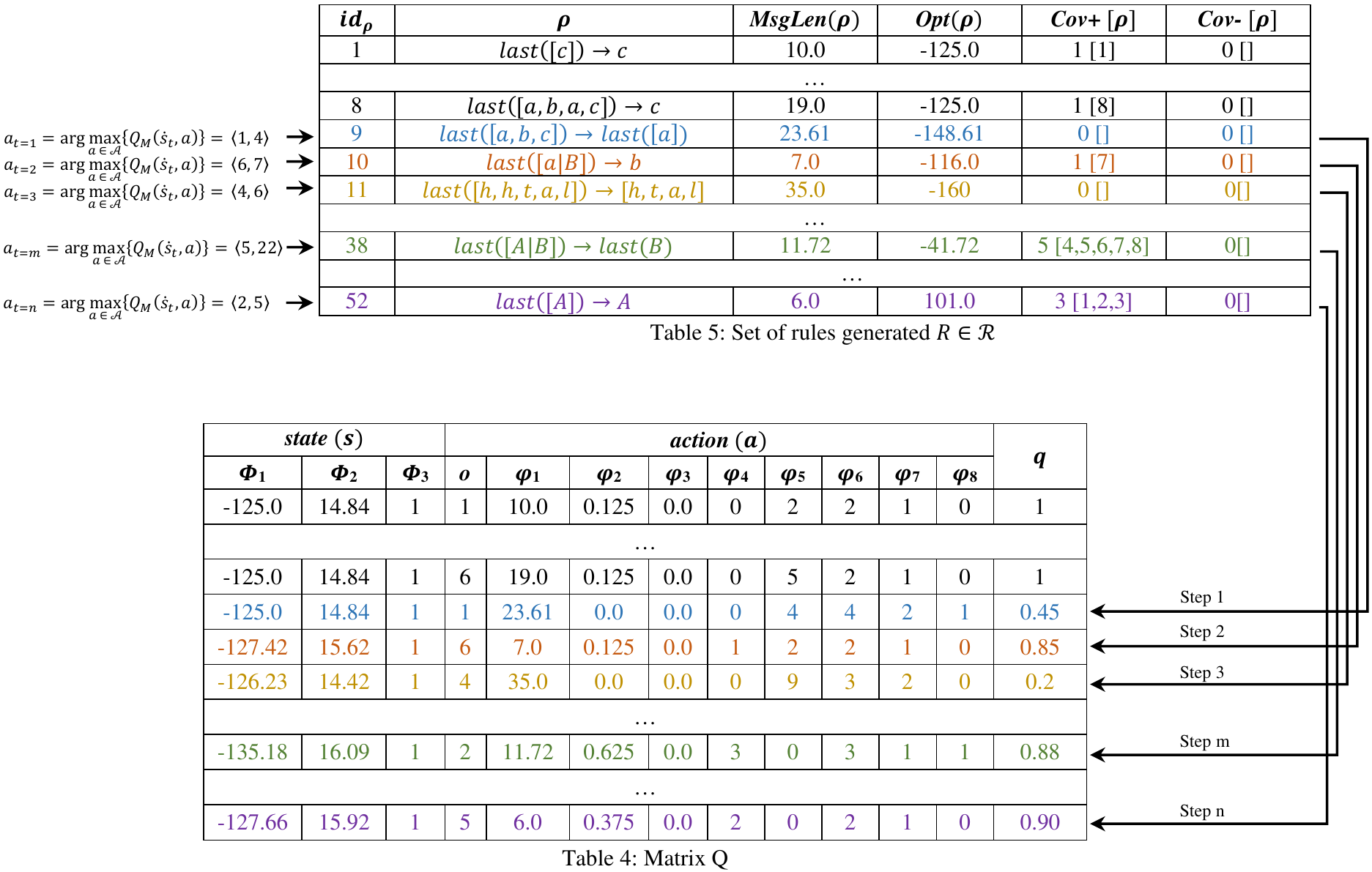}
	\caption{After initialising $R$ and Q-matrix, in each following step, \gerl selects an action updating the table Q and generating a new rule $\rho$ which is inserted into $R$ until the \emph{Stop Criterium} is reached}
	\label{fig:problemInstanceVerticalB}
\end{figure}

\section{Reusing past policies}\label{sec:reuse}

In this section, we describe how to reuse and apply the policies previously learnt. 
%

As we have seen in Section \ref{previous}, in other TL methods the knowledge is transferred in several ways (via modifying the learning algorithm, biasing the initial action-value function, etc.) 
 and,  if the source and the target task are very different, a mapping between actions and/or states is also needed. Instead of that, in \gerl the reuse of previous acquired knowledge is done in a different way, by reusing the values in the $Q$ table. 

The main reason why we can use past policies (the learned table $Q$) in order to accelerate the learning of different new tasks is due to the abstract representation of states and actions (the $\phi$ and $\varphi$ features of $\dot{\cal{S}}$ and $\dot{\cal{A}}$ respectively) which prevents the system from start from scratch. Actions which are successfully applied to certain states (from the previous task) are also applied when a similar (with similar features) new state is reached. Due to this abstract representation, how different the source and target task are does not matter.

\begin{figure}[htbp]

\centering
		\includegraphics[width=1.0\textwidth]{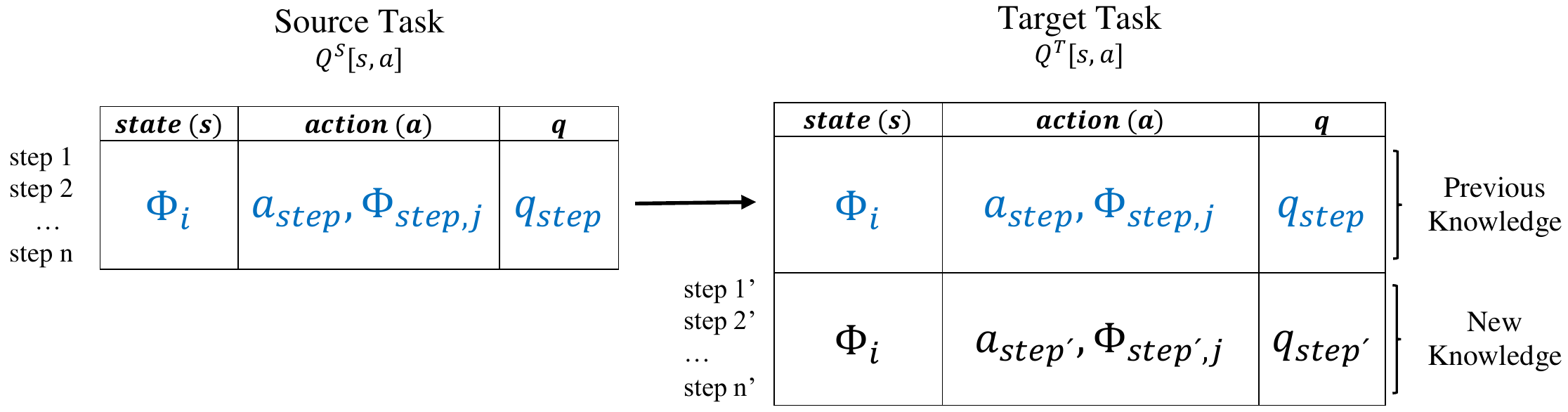}
	
	\caption{The knowledge transfer process in the \gerl system where $i \in 1..3$ and $j \in 1..8$ are, respectively, the set of features indexes used to describe the \emph{states} and \emph{rules}.}
	\label{fig:TLgErl}
	
\end{figure}

The knowledge transfer between two task (source $S$ and target $T$ respectively) is performed as follows: 
when \gerl reaches the solution of a given problem (or it executes a maximum number of steps), the table $Q^{S}$ (which has been filled in by the model $Q_{M}^{S}$ and equation \ref{upQ}) is copied and transferred to a new situation. 
Concretely, when \gerl learns the new task, $Q^{S}$ is used to train a new model $Q_{M}^{T}$\footnote{We do not transfer the $Q_{M}^{S}$ model  since it may not  have been retrained with the last information added to the table $Q^{S}$ (because of the periodicity of training, the generation of latest model may not match with the stopping criterion, so there may be a bunch of information in $Q$ which is not used to retrain $Q_M^{S}$).}. Therefore, $Q^{S}$ is used from the first learning step and it is afterwards updated with the new information assimilated by the model $Q_{M}^{T}$. Figure \ref{fig:TLgErl} briefly describes the process of reusing a  $Q^{S}$ table and how it becomes  a  new  $Q^{T}$ table. 
 
\subsection{An illustrative example of knowledge transfer}\label{TLexperiments}

In this section, we describe an experiment 
which illustrates the policy reuse strategy used in \gerl. 


We will use list processing problems as a structured prediction domain \cite{springerlink:10.1007/s10994-008-5079-1}  where not only the input is structured but also the output. For the first part of our study we have selected five different problems that 
use the Latin alphabet $\Sigma = \{a,b,\dots, z\}$ as a finite set of symbols and perform the following simple transformations between symbols:

\begin{enumerate}
	\item $d \ra c$: replaces \emph{``d''} by \emph{``c''}. Instances would look like this: \newline
$ trans([t,r,a,d,e]) \rightarrow [t,r,a,c,e] $
	\item $e \ra ing$: replaces \emph{``e''} by \emph{``ing''} located at the last position of a list. Instances would look like this:  $ trans([t,r,a,d,e]) \rightarrow [t,r,a,d,i,n,g] $
	\item $d \ra pez$: replaces \emph{``d''} by \emph{``pez''} located at any position of a list. Instances would look like this:  $ trans([t,r,a,d,e]) \rightarrow [t,r,a,p,e,z,e] $
	\item $Prefix_{over}$: adds the prefix \emph{``over''}. Instances would look like this: \newline
	$ trans([t,r,a,d,e]) \rightarrow [o,v,e,r,t,r,a,d,e] $
	\item $Suffix_{mark}$: adds the suffix \emph{``mark''}. Instances would look like this:\newline
	$ trans([t,r,a,d,e]) \rightarrow [t,r,a,d,e,m,a,r,k] $
\end{enumerate}

\noindent Since our aim now is to illustrate the ability of \gerl to transfer knowledge between different tasks, it is not so important to see which operators or functions are needed as we did in the previous sections, but how the system is able to accelerate the learning reusing knowledge. However, we will briefly describe the functions and operators needed to navigate the structure (lists) and apply local or global changes to it. It is easy to see that we need to provide the system with some functions (on the Background Knowledge $K$) in order to find differences between lists. 


\begin{itemize}
\item The function  $oneSust(List_1,List_2)$, is actually a composition of two functions. The first one is $difLists: list \times list \rightarrow (\Sigma \rightarrow \Sigma)$ which  takes two lists and goes through them comparing their symbols until a difference is returned as a replacement function (between symbols). The second function is the high-order function $map$ whose parameters are the replacement function returned by $difLists$  and $L_1$ (the first parameter of $oneSust$). Therefore,  the $oneSust$ function is defined as:

\begin{align*}
oneSust(List_1,List_2) \:\:\Rightarrow\:\:   map(difLists(List_1,List_2), List_1))
\end{align*}

\noindent An example of application of  this function may be:

{\small
\[\begin{array}{ll} 

oneSust([a,b,c],[d,b,c])     & \:\:\Rightarrow\:\: map(difLists([a,b,c],[d,b,c]), [a,b,c])) \:\:\Rightarrow\:\: 	\\ 
 map(a \rightarrow d, [a,b,c])   & \:\:\Rightarrow\:\:  [d,b,c] \\
\end{array}\]
}

\item The second function, $nSust(List_1,List_2)$,  is also a composition of functions similar to the previous one, but it uses the function $nDifLists: list \times list \rightarrow [(\Sigma \rightarrow \Sigma)]$ . This function, instead of returning the first simple replacement (one symbol by another),  it returns a list of as many replacement functions as possible matchings between the symbol of $L_1$ which is different to the corresponding symbol in $L_2$ and all consecutive sublists of $L_2$ (having a length greater than 1) that can be extracted from the difference. Then, the list of replacement functions is taken as the first parameter of a high order function $nmap$ that applies all of them to $L_1$ (the first parameter of $nSust$). Thus, the $nSust$ function is defined as:

\begin{align*}
		nSust(List_1,List_2) \:\:\Rightarrow\:\:   nmap(nDifLists(L_1,L_2), L_1)
\end{align*}

\noindent An example of application of  this function may be:

{\small
\[\begin{array}{ll} 

nSust([a,b,c],[a,d,e,c])     \:\:\Rightarrow\:\: nmap(nDifLists([a,b,c], [a,d,e,c]), [a,b,c]) \:\:\Rightarrow\:\: &	\\ 
 \lbrace
  \begin{array}{l}
     \  map(b\ra de,[a,b,c]) \:\:\Rightarrow\:\: [a,d,e,c]\\
     \  map(b\ra dec,[a,b,c]) \:\:\Rightarrow\:\: [a,d,e,c,c]
  \end{array}

\end{array}\]
}

\item The third and fourth background-knowledge functions we use are $add\-Prefix(List_1,List_2)$ and $addSuffix(List_1,List_2)$, which add $List_1$ as a prefix or a suffix (respectively) to the list obtained as the difference between $List_1$ and $List_2$:

\noindent 
\[addPrefix(List_1,List_2) \:\:\Rightarrow\:\:   List_1 ++ subtract(List_1,List_2)\]
\[addSuffix(List_1,List_2) \:\:\Rightarrow\:\:   subtract(List_1,List_2) ++ List_1\]

\noindent where $substract(List_1,List_2)$ is a $BiF$ (Built-in-Function) of the Erlang language that returns a new list $List_3$ which is a copy of  $List_1$, subjected to the following procedure: for each element in $List2$, its first occurrence in $List1$ is deleted.

Let us see an example: 

\[addPrefix([a,b,c],[a,b,c,z]) \:\:\Rightarrow\:\:   [a,b,c] ++ [z]\]
\[addSuffix([a,b,c],[z,a,b,c]) \:\:\Rightarrow\:\:   [z] ++ [a,b,c]\]

\end{itemize}


The above mentioned functions are used for defining the operators to be applied in the learning process:
\begin{itemize}

\item The first operator $op_1$  is defined as 
\[op_1\equiv meta\_replace(Rt_1,\- oneSust(L_1,R_1))\]

\noindent that replaces the $rhs$ of a rule by the application of the $oneSust$ function. 


\item The second operator $op_2$ is defined in a similar way but using the $nSust$ function:

\[op_2\equiv meta\_replace(\-Rt_1, \-nSust(L_1,R_1))\] 

\noindent Hence, it returns as many rules as possible replacement returns the $nSust$ function.

\item Two other operators, $op_3$ and $op_4$, are defined using the meta-operator $replace$ and the functions $addPrefix$ and $addSuffix $:

\[op_3\equiv meta\_replace(Rt_1, addPrefix(L_1,R_1))\]
\[op_4\equiv meta\_replace(Rt_1, \- addSuffix(L_1,R_1))\]

\noindent that will substitute the $rhs$ of a rule by the result of the functions.
\end{itemize}

\noindent Finally, we need a way of generalising the rules. That is performed by instantiating the meta-operator $replace$, $replace(Pos,V_{List})$,   with all the possible rule's positions $Pos$ where we can find a list ($L_1, Rt_1, Rt_{1.1}$ and $Rt_{1.2}$) as the first parameter  and a unique variable $V_{List}$ as second parameter.  As a result, we obtain four operators ($op_5, op_6, op_7$ and $op_8$), that is one operator for each position to be generalised. 

With these operators \gerl is able to solve any of the previous learning problems by simply applying a suitable sequence of  operators. For instance, given the instance $trans([a,b,c]) \rightarrow [a,d,c]$, the following sequence of actions lead to a possible solution:

\[\begin{array}{crll}
step_1: & \blue[op_1](trans([a,b,c]) \rightarrow [a,d,c]) & \:\:\Rightarrow\:\: & \\ & trans([a,b,c]) \rightarrow oneSust([a,b,c],[a,d,c]) & & \\
step_2: & \blue[one\_step\_rew](trans([a,b,c]) \rightarrow oneSust([a,b,c],[a,d,c])) & \:\:\Rightarrow\:\: & \\  & trans([a,b,c]) \rightarrow  map(b\ra d,[a,b,c]) & & \\
step_3: & \blue[op_8](trans([a,b,c]) \rightarrow map(b\ra d,[a,b,c])) & \:\:\Rightarrow\:\: & \\ & trans(V_{List}) \rightarrow map(b\ra d,[a,b,c]) & & \\
step_4: & \blue[op_5](trans(V_{List}) \rightarrow map(b \ra b,[a,b,c])) & \:\:\Rightarrow\:\: & \\ & trans(V_{List}) \rightarrow map(b\ra d,V_{List}) & &
\end{array}\]

\noindent where the latter rule $trans(V_{List}) \rightarrow map(b\ra d,V_{List}) $ is the solution.

Since we want to analyse the ability of the system to improve the learning process when reusing past policies, we will solve each of the previous problems separately and, next we will reuse the policy  from one problem to solve the rest  (including itself). 
The set of operators used consists of the user-defined operators we mentioned above and a small number of non-relevant operators 
we add to increase the difficulty of solving the problems. 
With all of this  the set of operators $O$ used has twenty operators. 
To make the experiments independent of the operator index, 
we will set up 5 random orders for them. 
Each problem has 20 positive instances and no negative ones. From each problem we will extract 5 random samples of ten positive instances in order to learn a policy from them with each of the five order of operators (5 problems $\times$ 5 samples $\times$ 5 operator orders $= $125 different experiments).

We show the aggregated means (in number of steps) of each sample and operator order without the reuse of previous policies (Table \ref{tab:woPCYs}) and with policy reuse (Table \ref{tab:wPCY}). To analyse whether the difference between solving the problem with and without  reusing the policy is significant, we performed the Wilcoxon signed-ranks test with a confidence level of $\alpha = 0.05$ and $N = 25$ (5 samples $\times$ 5 operators orders). 
 The results in bold means that the improvement is statistically significant. 
Interestingly, the results obtained reusing works for most combinations (for those combinations where the difference is not significant, it is still better in magnitude), including those cases where the problems have nothing to do and do not reuse any operator. This suggests that the abstract description of states and rules is beneficial even when the problems are not related. Actually, this gives support to the idea of a general system that can perform better as it sees more and more problems, one of the reasons why the reinforcement model and the abstract representations were conceived in \gerl.

\newlength{\mycolwd}
\settowidth{\mycolwd}{$Suffixx$}

\begin{table}[htbp]
{\scriptsize
  \centering

    \begin{tabular}{|C{\mycolwd}|C{\mycolwd} C{\mycolwd} C{\mycolwd} C{\mycolwd} C{\mycolwd}|}
		\hline

          & \textbf{$l \ra c$} &   \textbf{$e \ra ing$} & \textbf{$d \ra pez$} & \textbf{$Prefix_{over}$} & \textbf{$Suffix_{mark}$} \\\hline
					\textbf{Steps} & 108.68 & 76.76 & 74.24 & 61.28 & 62.28 \\\hline
        
            \end{tabular}%
   
		  \caption{Results not reusing previous policies  (average number of steps).}
  \label{tab:woPCYs}%

  \centering
  
    \begin{tabular}{|C{\mycolwd}|C{\mycolwd} C{\mycolwd} C{\mycolwd} C{\mycolwd} C{\mycolwd}|}
		\hline
          & \multicolumn{5}{c|}{\textbf{Problem}} \\\hline

     \textbf{PCY from}  & \textbf{$l \ra c$} &   \textbf{$e \ra ing$} & \textbf{$d \ra pez$} & \textbf{$Prefix_{over}$} & \textbf{$Suffix_{mark}$} \\\hline
    
    \textbf{$l \ra c$} & 		\textbf{65.68} & \textbf{58} & 70,64 & \textbf{48.84} & \textbf{49.12} \\ 
    \textbf{$e \ra ing$} &  \textbf{66.48} & \textbf{50.04} & \textbf{56.4} & \textbf{45.2} & \textbf{45.36} \\
    \textbf{$d \ra pez$} &  \textbf{56.36} & \textbf{49.6} & \textbf{57.32} & 52.24 & \textbf{45.84} \\
    \textbf{$Prefix_{over}$}& \textbf{58.8} & \textbf{48.96} & \textbf{60.6} & \textbf{43.8} & \textbf{46.88} \\
    \textbf{$Suffix_{mark}$} & 102,72 & \textbf{64.4} & 67.32 & 56.16 & \textbf{57.48} \\\hline
		\textbf{Average} & 70.01 & 54.2 & 62.46 & 49.25 & 48.94 \\\hline

    \end{tabular}%
		
		  \caption{Results reusing  policies (average number of steps).}
			
  \label{tab:wPCY}%
}
\end{table}%

\section{\gerl solving IQ Test Problems}\label{IQ}

The relation between artificial intelligence (AI) and psychometrics started several decades ago using IQ tests as targets problems for AI or as evaluation measures. While some computational models of psychometric test problems have been proposed all along the second half of the XXth century, in the first years of this century we have seen an increasing number of real systems being able to score well on specific IQ test tasks.

In this field, there are two distinct branches of thought followed by the AI researchers community: those who agree that IQ test are appropriate for evaluating the intelligence of machines or measuring the progress in AI \cite{Detterman2011}; and those who disagree \cite{iq,IQnotformachines}. One of the key issues in this debate is whether these problems have been addressed by specialised systems, which are able to solve one specific IQ-lile type of problem, but are unable to even attempt other problems. Also, most machine learning methods which are able to deal with large and complex datasets fail at these problems. In this section we will show how the same system, \gerl, can solve several IQ tests problems  just following the idea that most IQ test questions tended to follow a small number of patterns \cite{iq}.

Anyway, the application of the same system to this kind of problems may provide interesting information about the actual complexity of these problems on the one hand, and about the capabilities of \gerl on the other.

\subsection{Odd-one-out problems}

\begin{figure}
	\centering
		\includegraphics[width=1.0\textwidth]{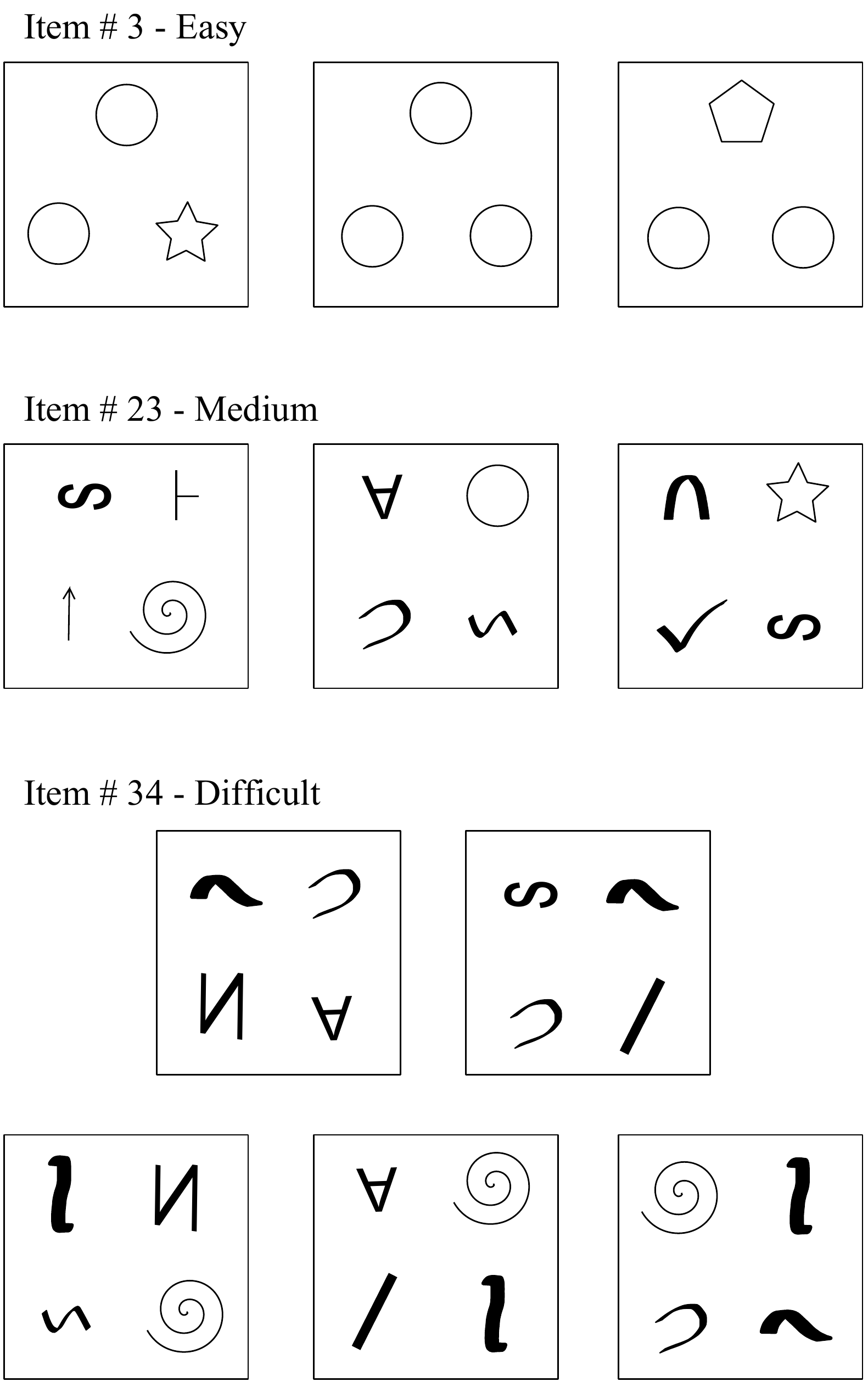}
	\caption{Examples 3, 23 and 34 in \cite{ruiz2011building} sorted by complexity (from easy to difficult).}
	\label{fig:ooo}
\end{figure}

Odd-one-out is a kind of problem where the goal is to detect the outlier in a set of objects. This is a fundamental task used in a wide variety of intelligence tests to evaluate human or animal intelligence. The odd-one-out problems were developed as a modern variant of some classical tests of fluid intelligence such as Raven's Progressive Matrices \cite{raven1992} and the Cattell's Culture Fair Intelligence Test \cite{feingold1940culture}. More precisely: given a set of items, the participant is asked to decide which one among them is most dissimilar from the
rest. Typically, the presented items can vary along one dimension (e.g., shape, size, quantity, \dots). The odd-one-out problems differ from Raven's and Cattell's intelligence tests as the problems are generated automatically using a complex set of algorithms \footnote{\url{http://www.cambridgebrainsciences.com/} provides scientifically proven tools for the assessment of cognitive function over the web.}. Due to this automatic generation, and the ability to generate many tens of thousands of novel problems. The participant cannot learn the answers to specific problems by rote learning. Consequently, the task is suitable for training reasoning abilities or taking many repeated measures. Figure \ref{fig:ooo} shows three examples of odd-one-out problems of increasing complexity.

Several approaches have been developed to solve odd-one-out problems. Visual approaches like  McGreggor \cite{McGreggor2011} that applies a novel analogical perspective using fractals in order to capture self-similarity and repetition at multiple scales; and Lovett \cite{lovett2008computational} who also uses an analogical generalisation using qualitative spatial representations with structure mapping. Finally, a different approach that assumes that some previous feature transformation has been made is followed by Ruiz \cite{ruiz2011building}. He uses the so-called ``Ruiz-Absolute Scale of Complexity Management'' (R-ASCM) where the objects in each item of an example are coded (``discretised'') as strings (by using letters of the Roman alphabet).


\begin{table}[htbp]
\centering
{\sffamily\small
\begin{tabular}{cccccc} 
\hline
Item 				      & Set1				& Set 2			&Set 3			&Set 4			&Set 5 \\ \hline

1       					& AAA        	&	AAA				& \cellcolor{grey}{ABB}				&						&    \\
2       					& AAA        	&	AAA				& \cellcolor{grey}{BCD}				&						&    \\
3       					& \cellcolor{grey}{AAA}        	&	AAB				& AAC				&						&    \\
4       					& \cellcolor{grey}{AAA}        	&	ABB				& ABB				&						&    \\
5       					& AAA        	&	BBB				& \cellcolor{grey}{ABC}				&						&    \\

6       					& \cellcolor{grey}{AAA}       	&	BCD				& EFG				&						&    \\
7       					& \cellcolor{grey}{AAA}&	BBC				& CCB				&						&    \\
8       					& AAB        	&	AAB				& \cellcolor{grey}{ABC}				&						&    \\
9       					& AAB        	&	AAC				& \cellcolor{grey}{DEF}				&						&    \\
10       					& AAB        	&	ABB				& \cellcolor{grey}{EFG}				&						&    \\

11       					& ABC        	&	ABC				& \cellcolor{grey}{ABD}				&						&    \\
12       					& AAB        	&	ABB				& \cellcolor{grey}{ABC}				&						&    \\
13       					& ABC        	&	ADE				& \cellcolor{grey}{FGH}				&						&    \\
14       					& \cellcolor{grey}{AAAA}     	 	& BBDE			& CCFG			&						&    \\
15       					& \cellcolor{grey}{AAAA}        &	AABB			& AACC			&						&    \\

16       					& \cellcolor{grey}{AAAD}        &	BBEF			& CCGH			&						&    \\
17       					& AABB        &	AABB			& \cellcolor{grey}{ABCD}			&						&    \\
18       					& \cellcolor{grey}{AABC}        &	AACD			& ABCD		&						&    \\
19       					& AAAB       	&	BBBD			& \cellcolor{grey}{CCCE}			&						&    \\
20       					& ABCD       	&	ABCD			& \cellcolor{grey}{ABCE}			&						&    \\

21       					& ABCD        &	ABCE			& \cellcolor{grey}{ABFG}			&						&    \\
22       					& AABC        &	BBAC			& \cellcolor{grey}{CCAF}			&						&    \\
23       					& ABCD       	&	AEFG			& \cellcolor{grey}{HIJK}			&						&    \\
24       					& AAAA       	&	AAAA			& BBBB			& BBBB		  & \cellcolor{grey}{CCCC}    \\
25       					& AAAD        &	AAAE			& BBBF			& BBBG			& \cellcolor{grey}{CCCH}    \\

26       					& AABB        &	BBCC			& AADD			& DDCC			& \cellcolor{grey}{EEFF}   \\
27       					& AAEF        &	BBGH			& CCIJ			& DDKL			& \cellcolor{grey}{ABCD}   \\
28       					& AAAE        &	BBBF			& CCGH			& DDIJ			& \cellcolor{grey}{ABCD}   \\
29       					& AAAE       	&	BBBF			& CCGH			& DDIJ			& \cellcolor{grey}{AABB}   \\
30       					& AAAB       	&	BBBF			& CCGH			& DDIJ			& \cellcolor{grey}{AABB}    \\

31       					& AABB        &	BBCC			& AADD			& DDCC			& \cellcolor{grey}{AAEE}   \\
32       					& ABCD       	&	BCDE			& CDEF			&	DEFG			& \cellcolor{grey}{FGAB}    \\
33       					& ACDE       	&	AFGH			& BIJK			&	BLMN			& \cellcolor{grey}{OPQR}   \\
34       					& ABEF       	&	ABGH			& CDEG			&	CDFH			& \cellcolor{grey}{ABCD}   \\
35       					& ACDE       	&	AFGH			& BIJK			&	BLMN			& \cellcolor{grey}{ABOP}   \\\hline

\end{tabular}
}
\caption{35 examples of R-ASCM abstract representation (solutions in grey) from Ruiz \cite{ruiz2011building}.}
\label{fig:tableRuiz}
\end{table}

\subsubsection{Using \gerl}

In order to understand how odd-one-out problems work, it is worth studying how to abstract them. To avoid starting from the scratch, our work is based on the abstract representation (R-ASCM)  followed by Ruiz \cite{ruiz2011building}.

The first step is to code the examples as equations,  in order to be correctly handled by \gerl. We use the R-ASCM coding where, for instance, an example composed by two circles and a square is represented as $(A, A, B)$ where $B$ is the odd-one out. 
This simple and compressed data abstraction makes sense as the research in \cite{Chater03simplicity:a} reveals that the goal of the cognitive systems is to compress data: the choices between patterns depend on the degree of compression such patterns provide (e.g., the higher the compression, the more compatible the patterns are with a finite body of data). Table \ref{fig:tableRuiz} shows the 35 examples that we will try to solve with the R-ASCM coding. 

More concretely, we represent an example (set of items) as a list of lists. The $rhs$ of the examples are numbers indicating which item is the odd-one-out. We will evaluate our system with the items in Figure \ref{fig:tableRuiz}. For instance, the example number 3 in Table \ref{fig:tableRuiz} is represented as: 

$$ ooo([[A,A,A], [A,A,B], [A,A,C]])\rightarrow 1.$$

Next we need to define appropriate operators, both to navigate the structure and to apply local or global changes to the rules. First of all, we can take advantage of the higher order function $map$ that Erlang provides, in order to apply functions over lists. Hence, we will use the meta-operator $meta\_replace$ for defining a general operator $op_1$

\[\begin{array}{cl}
\blue[op_1]\equiv & meta\_replace(Rt_1, map(\_,L_1))) \\
\end{array}\]

\noindent which will substitute the \emph{Right} part ($Rt_1$ position) of an input rule  by the expression in that the higher order function $map$ applies a function (which we represent by the anonymous variable \_) to the whole input list located in position $L_1$. 


As in \cite{ruiz2011building}, we will use similarity functions in order to compute distances between lists. With the aim of avoiding the need of recoding some misclassified items, we can provide the system with different similarity or distance functions between lists (as functions in the Background Knowledge). For instance, in addition to the Hamming Distance as used in \cite{ruiz2011building}, we implement a simple distance function that only counts the number of different objects inside an item ($diffObj$). For instance, if we have the example [[A,A,A], [A,A,B], [A,A,C]], the previous simple distance gives (1,2,2), so the odd-one out is the first item, which coincides with the solution. Since these functions have to be applied to a list, it can be done by using them as the first parameter of $map$ function. Therefore, we define operators $op_2$ and $op_3$ as

\[\begin{array}{cl}
\blue[op_2]\equiv & meta\_replace(R_{1.1}, hamming(L_1))\\
\blue[op_3]\equiv & meta\_replace(R_{1.1}, diffObj(L_1))\\
\end{array}\]

Although we already have operators to calculate similarities between lists, we need a way to select which the different item is with respect to the others. Since the $map$ function returns its input list transformed by the function, we can use it to apply a function that selects which item is the different one. This function is defined in the Background Knowledge ($distinct$) and we define the operator $op_4$ as 

\[\begin{array}{cl}
\blue[op_4]\equiv & meta\_replace( Rt, distinct(Rt))\\
\end{array}\]

Finally, we need a way of generalising the examples. That is performed by the meta-operator $replace$ instantiated with all possible positions where we can find a list ($L_1$, $Rt_{1.2}$ and $Rt_{1.1.2}$) and a variable as a second parameter ($V_{lists}$). This gives three generalisation operators ($op_5$, $op_6$ and $op_7$).

\subsubsection{Results}

Figure \ref{fig:tableRuiz} shows the 35 examples used to test our system in order to allow a comparative with the other approaches. Regarding the results, we only compare with Ruiz \cite{ruiz2011building} due to Lovett \cite{lovett2008computational} has a lower level of achievement (similar to children's achievement) and McGreggor \cite{McGreggor2011} does not makes any comparison.  Ruiz defines a 2-step odd-one-out clustering algorithm that compare each item with the rest trying to find the odd-one-out. The reason why the algorithm is named ``2-step'' is because it needs a second re-coding (after using the abstract representation (R-ASCM)) of the items in order to achieve the maximum accuracy. In the first step its algorithm is applied and 28 of 35 (80\%) examples are solved. The second step consists in taking the examples misclassified (he knows the classes of the test set), recoding them and launching the algorithm again. In this second step 6 of the 7 remaining examples are solved, however, this result can not be added to the previous one for obvious reasons.


As we have mentioned in the previous section, our approach is more general than Ruiz is in the sense that we do not have to recode examples misclassified because we use two different distance functions. Table \ref{tab:ooo1} shows the two better rules founded by \gerl (which have the highest optimality).

\begin{table}[htdp]
\begin{center}
{\small
\begin{tabular}{ccc}\hline 
Rank. &  Rule                                                               & Cov.     \\ \hline 

1			&  $ooo(V_{lists}) \Rightarrow distinct(map(hamming,V_{lists}))$       & 28/35     \\ 
2			&  $ooo(V_{lists}) \Rightarrow distinct(map(diffObj,V_{lists}))$       & 17/35      \\

\hline\end{tabular}
}
\end{center}
\caption{Best two rules founded by \gerl. }
\label{tab:ooo1}
\end{table}%


\noindent taking the first rule, 28 of 35 (80\%) examples are solved (being on a par with an average human adult), the same result that Ruiz's algorithm achieves (it can not be taken into account the second re-coding). Regarding the second best rule learned, it solves 17 of 35 examples. Note that some examples are covered by both rules. Table \ref{tab:ooo2} shows the classification results. Example number 31 is not covered by any rule because it exhibits other mathematical properties, not captured by the distance functions used.

\begin{table}[htdp]
\begin{center}

\begin{tabular}{c}

		\includegraphics[width=1.00\textwidth]{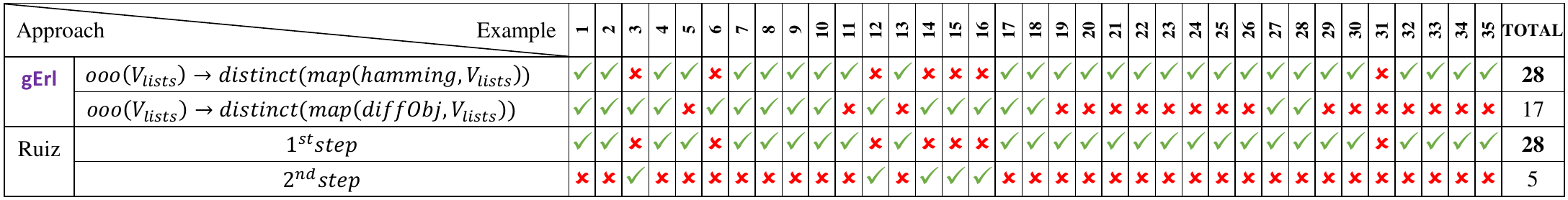}

\end{tabular}

\end{center}
\caption{Classification results for both \gerl and Ruiz approach \cite{ruiz2011building}}
\label{tab:ooo2}
\end{table}%

\subsection{Raven's IQ Tests}

\begin{figure}
	\centering
		\includegraphics[width=0.5\textwidth]{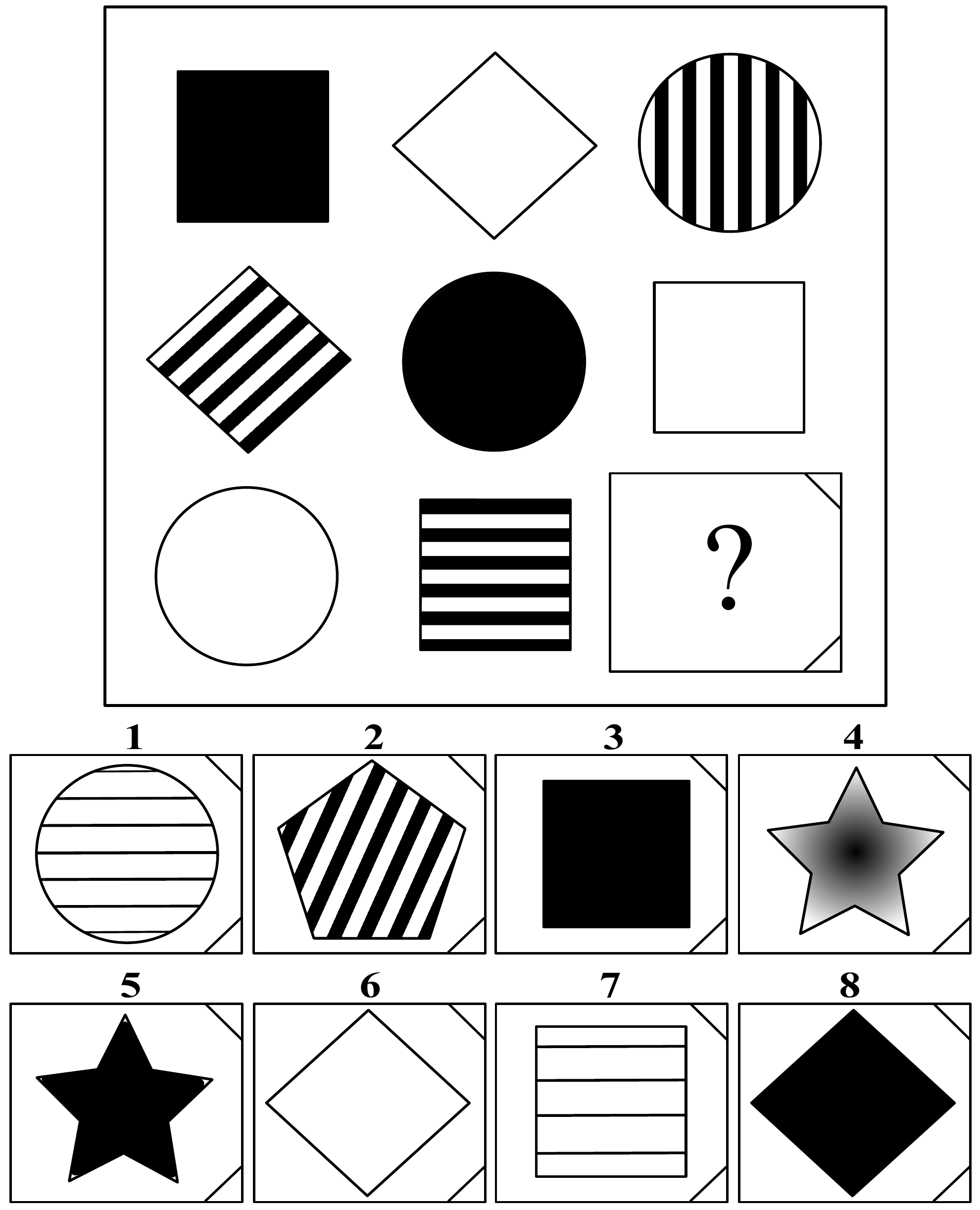}
	\caption{A geometrical reasoning problem. The solution is no. 8. For copyright reasons, the illustrations in this paper do not depict original RPM problems, but constructed equivalents}
	\label{fig:raven}
\end{figure}

This test was designed to measure average intelligence. Raven's intelligence test consists of a ($3 \times 3$) matrix. Eight possible choices ares displayed at the bottom (Figure \ref{fig:raven}). A figure is placed at each of the nine positions except the bottom-right one which is empty. There is a logical relation between the figures, which can be seen either horizontally (rows) or vertically (columns). The goal is to find this relation and choose a suitable figure among the eight figures proposed as solutions for the gap. Consider the following problem (Figure \ref{fig:raven}), which is an example of one kind of Raven's Progressive Matrices. The task of the reasoner is to guess the hidden geometrical function that has been followed to generate the example and apply it to infer the solution. 

There are currently three published versions of the Raven's Progressive Matrices (RPM) \cite{raven1992}: the original Standard Progressive Matrices (SPM), the Advanced Progressive Matrices (APM), developed as a more difficult test than the SPM for individuals in high IQ ranges, and the Colored Progressive Matrices (CPM), intended as a simpler test than the SPM to be used with children, the elderly, or other individuals falling into lower IQ ranges \cite{mccallum2003handbook}.

Several approaches have been developed to solve this kind of IQ tests. There are two main families: those works that try to solve RPM in an AI-like fashion \cite{evans1964program,mcgreggor2010fractal,Strannegard2013progressive}, and those that try to solve them similarly to humans \cite{carpenter1990one,lovett2007analogy,lovett2010structure,ragni2012solving}. We focus this section on the latter and with them we will compare. Carpenter \cite{carpenter1990one} is probably the oldest example in this family, with the goal of better understand human intelligence and the nature of the tests. He analysed the rules needed to solve the APM and produced a pair of computer simulation models called $FAIRAVEN$ and $BETTERAVEN$ that performed like the median or best college students in the sample, respectively. Lovett \cite{lovett2007analogy,lovett2010structure} developed and RPM solver based on Carpenter's work using an analogical reasoning strategy. Ragni and Neubert \cite{ragni2012solving} presented another system for Raven's Progressive Matrices, implemented in the cognitive architecture ACT-R \cite{citeulike:3371219}, which consists in a production rule system distributed in layers where the knowledge is represented by chunks (n-tuples) coding the information about objects and their relative position. Briefly explained, the Ragni and Neubert's system requires the identification of five relational rules (described below), as Carpenter did, and attributes to describe the information stored about the images: shape, size, number of sides, width, height, colour, rotation, position and quantity. 
Although the results were not compared to humans, they are compared to Carpenter's BETTERAVEN, with similar (or slightly better) score, so we estimate the results to be around human average.

\subsubsection{Using \gerl}

We will focus only in learning the SPM, which consist of 5 sets with 12 items in each set - 60 items in total. A and B test sets are for children under 14 years of age, and C, D, and E test sets are for over 14 years old persons. Each set starts with a problem which is, as far as possible, self-evident and becomes progressively more difficult. 
For the purposes of this work we have used the subsets C to E, trying to build the solution rather than choose among the collection of possible solutions given.

In order to enable the \gerl system to learn to solve RPM, a feature-based coding similar to that of Ragni and Neubert \cite{ragni2012solving} is needed. Also, a list-based coding is used to represent cells, rows, and matrices: Every figure inside a cell is abstractly represented as a tuple of features: $$\left\langle shape, size, quantity, posi\allowbreak ti\allowbreak on, ty\-pe \right\rangle$$ Every cell is represented as a list of figures. For instance, the way to describe the top-left cell in Figure \ref{fig:raven} which contains a simple big black square is:

\[ \big[\left\langle  squared,big,1,none,black \right\rangle_{figure_1}\big]_{cell_{1,1}} \]

\noindent Every row is represented as a lists of cells, and, finally, every Raven's matrix as a list of rows.

\begin{figure}
	\centering
		\includegraphics[width=1.0\textwidth]{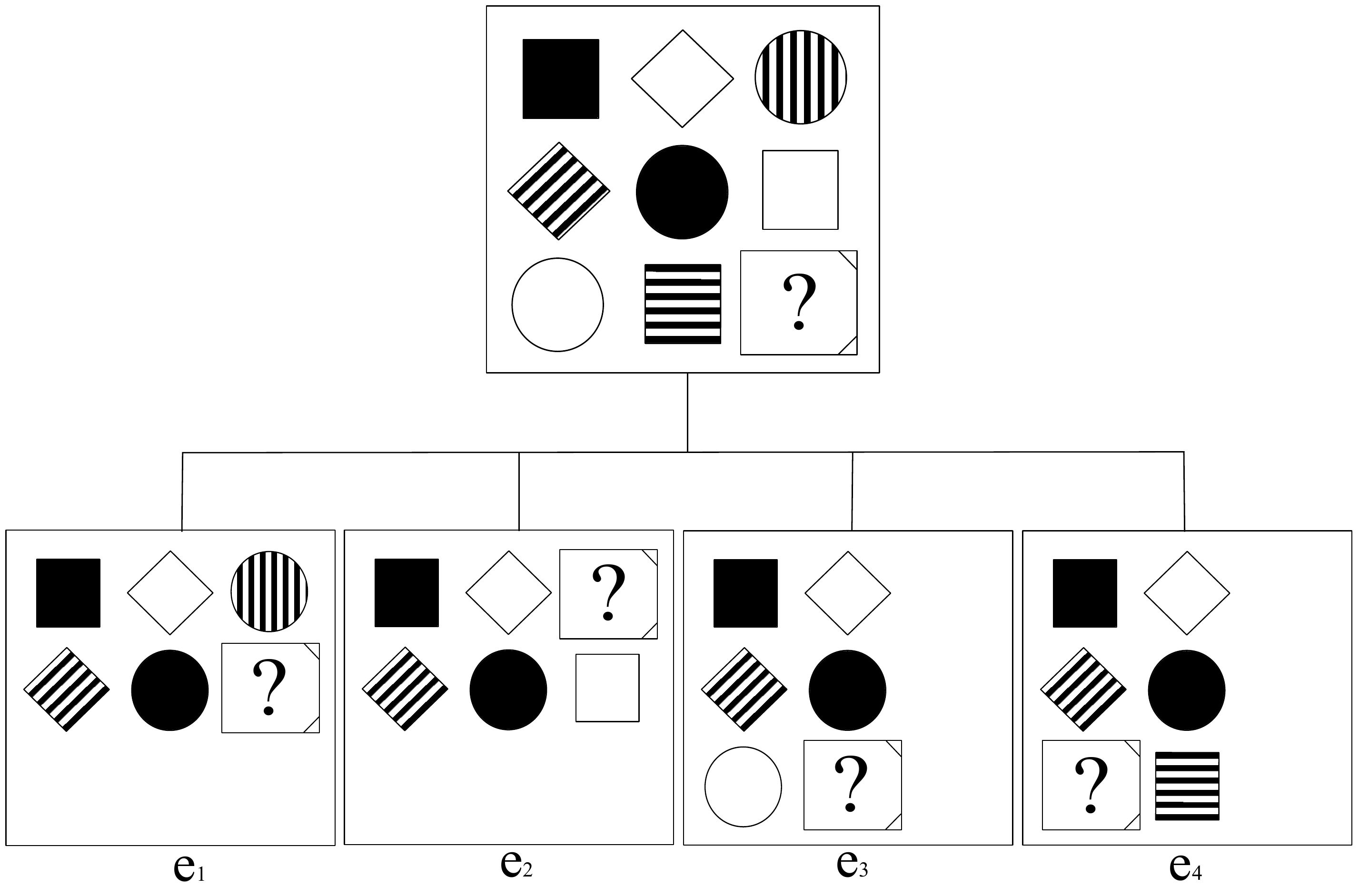}
	\caption{Raven's matrix decomposition example.}
	\label{fig:raven2}
\end{figure}

Since every single Raven's matrix is a problem itself (each matrix shows a different pattern), we need a way to generate several instances in order to make learning possible, by taking the most information from each matrix. To do that, each matrix is decomposed in several sub-matrices (the number depends on the problem) as we can see in Figure \ref{fig:raven2}. The last row/column cannot be used since they contain the gap to be filled in. However, they will be used to create up to 16 test examples replacing the gap by each of the eight possible solutions. In this way, the problem will be successfully learned if the test example covered by the solution obtained by the system coincides with the right answer.

Taking the instance $e_1$ from Figure \ref{fig:raven2} as an example, it will have the following representation:

\[e_1: \;\; raven(\bigg[\Big[\big[\left\langle square,big,1,none,black\right\rangle\big],\big[\left\langle diamond,big,1,none,white\right\rangle\big],\]
\[\big[\left\langle circle,big,1,none,striped\right\rangle\big]\Big],  \]
\[\Big[\big[\left\langle diamond,big,1,none,striped\right\rangle\big], \big[\left\langle circle,big,1,none,black\right\rangle\big]\Big]\bigg]) \rightarrow\]
\[\big[\left\langle square,big,1,none,white\right\rangle\big].\]

\begin{figure}[htbp]
	\centering
		\includegraphics[width=1.00\textwidth]{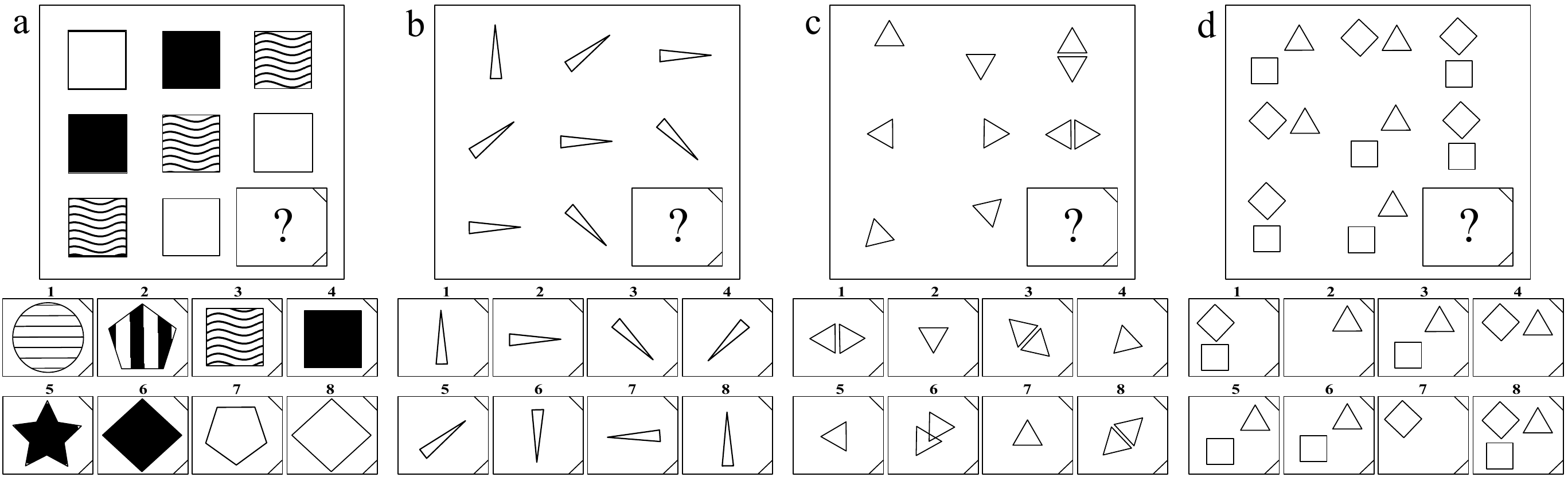}
		\caption{Problems illustrating the different rules (extracted from \cite{ragni2012solving}). In \ref{fig:Ravensrules}a the rule \emph{distribution of three values} is required. The solution is no. 4. Problem \ref{fig:Ravensrules}b requires the rule \emph{quantitative pairwise progression}. The solution is no. 6. In problem \ref{fig:Ravensrules}c the rule \emph{figure addition} is required. The solution is no. 8. Problem \ref{fig:Ravensrules}d the rule \emph{distribution of two values} is required. The solution is no. 4.}
	\label{fig:Ravensrules}
\end{figure}

To solve the problems of the Progressive Matrices, it is necessary to describe the relations between the objects in a row or column.  Carpenter \cite{carpenter1990one} identified five relations:

\begin{enumerate}
	\item {\bf{Relation 1: Constant in a row / Identity}}: If the value of a particular attribute of different objects remains constant in a row, put this attribute value in the solution cell.
	\item {\bf{Relation 2: Distribution of three values / Distribution of three entities}}: If an attribute of the objects in a row differs (3 different values) and the same values of this attribute occur in every row, put the missing value of this attribute in the solution cell. In Figure \ref{fig:Ravensrules}a, the attribute \emph{type} differs.
	\item {\bf{Relation 3: Quantitative pairwise progression / Numeric Progression}}: If the values of an attribute are in an increasing or decreasing sequence in all rows, put the following value into the solution cell. This rule is used in Figure \ref{fig:Ravensrules}b, where the object's position attribute increases by 45 degrees in each cell.
	\item {\bf{Relation 4: Figure addition / Binary OR}}: If all objects of the third column appear in their respective rows in the first two columns, put an addition of the objects of the two first cells into the solution cell. An example is depicted in Figure \ref{fig:Ravensrules}c.
	\item {\bf{Relation 5: Distribution of two values / Binary XOR}}: if there are exactly two equal values and one differing value of an attribute in  the rows, put an $XOR$ of the objects of the two first cells in the third row into the final cell. This rule is necessary to solve the problem \ref{fig:Ravensrules}d.
\end{enumerate}

To solve the selected 36 matrices from SPM, we need a way to apply the previous relations (which are implemented as functions located in the Background Knowledge) to the different attributes of the figures. 
The meta-operator $meta\_replace$ fits perfectly to this end by defining operators that apply the above-mentioned relations (implemented as functions) to the rules. These operators are defined following that scheme: $$meta\_replace(Position(Att), AddhocRule(Att))$$ where $AddhocRule$ corresponds to one of the five previous relations applied to an attribute $Att$  ($shape$, $size$, $quantity$, $position$ and $type$) where $Position(Att)$ returns the position of the descriptive attribute $Att$.

We will obtain as many operators as kinds of attributes multiplied by the number of relations (5 attributes $\times$ 5 relations, 25 operators). For instance, the operators that apply the first relation  ($identity$) to the five attributes will be defined as:

\[\begin{array}{cl}
\blue[op_1]\equiv & meta\_replace(Position(shape), identity(shape)) \\
\blue[op_2]\equiv & meta\_replace(Position(size), identity(size)) \\
\blue[op_3]\equiv & meta\_replace(Position(quantity), identity(quantity)) \\
\blue[op_4]\equiv & meta\_replace(Position(position), identity(position)) \\
\blue[op_5]\equiv & meta\_replace(Position(type), identity(type)) \\
\end{array}\]

Below we can see an example where the system applies the operator \blue[$op_3$] to a rule:
{\scriptsize
$$\blue[op_3](raven(V_{Matrix})\rightarrow \big[\left\langle distrib3val(shape), identity(Size), 1,none, distrib3val(type) \right\rangle\big]) \:\:\Rightarrow\:\: \\$$
$$raven(V_{Matrix}) \rightarrow \big[\left\langle distrib3val(shape), identity(size), identity(quantity), none, distrib3val(type)\right\rangle\big]$$
}

\noindent where $distrib3val$ corresponds with the function which implements the third previous relations and where $V_{Matrix}$ is the generalisation of an item, which suggests that we need also generalisation operators to generalise the input lists (in order to create a more general rule): $meta\_replace(L_1, V_{Matrix})$. If we apply the previous rule learned by the system to the test row/column (Figure \ref{fig:raven}, it returns the following cell as solution:

\[ \big[\left\langle diamond, big, 1,none,black\right\rangle\big] \]

\noindent which covers the correct solution (no. 8).

\subsubsection{Results}

\begin{table}
{\scriptsize{

\centering
{\sffamily\tiny
\begin{tabular}{clccc}
\hline 
Id														& Solution																																																																	& Steps 			& $E^+$				& $o$	\\
\hline

{\bf{25}} 										& $ravenPM(V) \rightarrow [\left\langle  identity(shape), none, none, none, none \right\rangle]	$																							&  37					& 3						  &	2   \\
{\bf{26}} 										& $ravenPM(V) \rightarrow [\left\langle  identity(shape), progressive(size), none, none, none \right\rangle]	$																&  99					& 4						  &	3   \\
{\bf{27}} 										& $ravenPM(V) \rightarrow [\left\langle  identity(shape), progressive(size), none, none, none \right\rangle]	$																&  99					& 4						  &	3   \\
{\bf{28}} 										& $ravenPM(V) \rightarrow [\left\langle  identity(shape), progressive(size), none, none, none \right\rangle]	$																&  111				& 4						  &	3   \\
{\bf{29}} 										& $ravenPM(V) \rightarrow [\langle  identity(shape), none, progressive(quantity),$																												&  131				& 4						  &	4   \\
															&																																											$progressive(position), none \rangle]	$						&  						& 					    &	    \\
{\bf{30}} 										& $ravenPM(V) \rightarrow [\left\langle  identity(shape), progressive(size), none, none, none \right\rangle]	$																&  88					& 4			  		  &	3   \\
{\bf{31}} 										& $ravenPM(V) \rightarrow [\left\langle  identity(shape), none, none, progressive(position), none \right\rangle]	$														&  81					& 4						  &	3   \\
{\bf{32}} 										& $ravenPM(V) \rightarrow [\left\langle  identity(shape), none, progressive(quantity), none, none \right\rangle]	$			   										&  79					& 4						  &	3   \\
{\bf{33}} 										& $ravenPM(V) \rightarrow [\left\langle  identity(shape), none, none, progressive(position), none \right\rangle]	$			   										&  91					& 4						  &	3   \\
{\bf{34}} 										& $ravenPM(V) \rightarrow [\left\langle  identity(shape), none, none, progressive(position), none \right\rangle]	$			   										&  91					& 4						  &	3   \\
{\bf{35}} 										& $ravenPM(V) \rightarrow [\left\langle  identity(shape), none, progressive(quantity), none, none \right\rangle]	$			   										&  81					& 4						  &	3   \\
{\bf{36}} 										& $ravenPM(V) \rightarrow [\left\langle  identity(shape), none, none, progressive(position), none \right\rangle]	$														&  83					& 4						  &	3   \\
{\bf{37}} 										& $ravenPM(V) \rightarrow [\left\langle  identity(shape), none, none, none, identity(type) \right\rangle]	$			   														&  75					& 4						  &	3   \\
{\bf{38}} 										& $ravenPM(V) \rightarrow [\left\langle  distrib3val(shape), none, none, none, none \right\rangle]	$			   																	&  69					& 4						  &	2   \\
{\bf{39}} 										& $ravenPM(V) \rightarrow [\left\langle  distrib3val(shape), none, none, none, none \right\rangle]	$			   																	&  71					& 4						  &	2   \\
{\bf{40}} 										& $ravenPM(V) \rightarrow [\left\langle  identity(shape), none, none, none, distrib3val(type) \right\rangle]	$			   												&  94					& 6						  &	3   \\
{\bf{41}} 										& $ravenPM(V) \rightarrow [\left\langle  identity(shape), none, none, none, distrib3val(type) \right\rangle]	$																&  96					& 6						  &	3   \\
{\bf{42}} 										& $ravenPM(V) \rightarrow [\left\langle  identity(shape), none, none, none, distrib3val(type) \right\rangle]	$		   													&  93					& 6						  &	3   \\
{\bf{43}} 										& $ravenPM(V) \rightarrow [\left\langle  distrib3val(shape), none, none, none, distrib3val(type) \right\rangle]	$		   												&  106				& 6						  &	3   \\
{\bf{44}} 										& $ravenPM(V) \rightarrow [\left\langle  distrib3val(shape), none, none, none, distrib3val(type) \right\rangle]	$		   		   									&  91					& 6						  &	3   \\
{\bf{45}} 										& $ravenPM(V) \rightarrow [\left\langle  distrib3val(shape), none, none, none, distrib3val(type) \right\rangle]	$															&  104				& 6						  &	3   \\
{\bf{46}} 										& $ravenPM(V) \rightarrow [\left\langle  identity(shape), none, none, none, distrib3val(type) \right\rangle]	$																&  93					& 6						  &	3   \\
{\bf{47}} 										& $ravenPM(V) \rightarrow [\langle  identity(shape), none, distrib3val(quantity), none,$																									&  146				& 6						  &	4   \\
															& 																																													$distrib3val(type) \rangle]	$								&  						& 						  &	   \\
{\bf{48}} 										& $ravenPM(V) \rightarrow [\left\langle  distrib3val(shape), none, none, none, distrib3val(type) \right\rangle]	$		   												&  106				& 6						  &	3   \\
{\bf{49}} 										& $ravenPM(V) \rightarrow [\left\langle  addition(shape), none, none, none, none	\right\rangle]	$		   		   																&  61					& 4						  &	2   \\
{\bf{50}} 										& $ravenPM(V) \rightarrow [\left\langle  addition(shape), none, none, none, none \right\rangle]	$																							&  55					& 4						  &	2   \\
{\bf{51}} 										& $ravenPM(V) \rightarrow [\left\langle  addition(shape), none, none, none, identity(type) \right\rangle]	$																		&  99					& 6						  &	3   \\
{\bf{52}} 										& $ravenPM(V) \rightarrow [\left\langle  distrib2val(shape), none, none, none, none \right\rangle]	$																					&  63					& 4						  &	2   \\
{\bf{53}} 										& $ravenPM(V) \rightarrow [\left\langle  distrib2val(shape), none, none, none, none \right\rangle]	$									   											&  60					& 4						  &	2   \\
{\bf{54}} 										& $ravenPM(V) \rightarrow [\left\langle  distrib2val(shape), none, none, none, none \right\rangle]	$		   		   															&  61					& 4						  &	2   \\
{\bf{55}} 										& $ravenPM(V) \rightarrow [\left\langle  distrib2val(shape), none, none, none, none \right\rangle]	$																					&  77					& 4						  &	2   \\
{\bf{56}} 										& $ravenPM(V) \rightarrow [\left\langle  distrib2val(shape), none, none, none, none \right\rangle]	$																					&  99					& 4						  &	3   \\
{\bf{57}} 										& $ravenPM(V) \rightarrow [\left\langle  distrib2val(shape), none, none, none, distrib3val(type) \right\rangle]	$												 			&  10					& 4						  &	3   \\
{\bf{58}} 										& $ravenPM(V) \rightarrow [\left\langle  distrib2val(shape), none, none, none, none \right\rangle]	$							   													&  60					& 4						  &	2   \\
{\bf{59}} 										& $ravenPM(V) \rightarrow [\left\langle  distrib2val(shape), none, none, none, none \right\rangle]	$		   		   															&  65					& 4						  &	2   \\

\hline

\end{tabular}
}
\caption{Solutions returned and steps needed by \gerl to learn the Raven's Standard Progressive Matrix \cite{SimonKotovsky1963}.}
\label{tab:resSPM}
}}
\end{table}

As we have said, our system has been tested on different sets which are functionally equivalent to the sets C through E of the Standard Progressive Matrices. The data (matrices pictures) was collected from \cite{RavenWeb}.

Our motivation was to test \gerl trying to solve geometrical reasoning problems but in a simpler way than \cite{carpenter1990one,lovett2007analogy,lovett2010structure,ragni2012solving} did, just using a feature representation of the examples and applying the well-know five \emph{relations} as functions enquiring about specific attributes. \gerl is able to solve 35 out of the 36 problems (12 of 12 of sets C and D, and 11 of 12 of set E). Compared to Lovett \cite{lovett2007analogy,lovett2010structure}, they tested on SPM sections B to E and they only report global results: 44/48 correct answers, but it is assumed that none of their missed problems are part of section B, therefore they obtained a score of 32/36 on sections C to E. With respect to Ragni \cite{ragni2012solving}, our results are in line with theirs (35 out of 36) for SPM matrices.  In their work, Ragni compared to Carpenter's BETTERAVEN \cite{carpenter1990one}, with similar (or slightly better) score, so, we can estimate our results to be on the human average. Table \ref{tab:resSPM} shows the solution learned by \gerl.

Although the results can not be compared to humans, we converted this score to 59/60 on the overall test, as individuals who performed this well on the later sections typically got a perfect score in sections A and B \cite[table SPM2]{raven12court}.  
A score of 59/60 is in the $95^{th}$ percentile for American adults (IQ: ~140), according to the 1993 norms \cite{raven1996manual}, above the human average.


\subsection{Thurstone letter series completion problems}

The goal of this sort of problems, commonly found in intelligence tests, is to guess the following letter in a series (see Table \ref{fig:letter}). It was used by Thurstone \cite{thurstone1941factorial} in their studies of intelligence. More recently,  performance on this task has been investigated in computer simulation studies. Simon and Kotovsky \cite{SimonKotovsky1963,Kotovsky1973399} provided what is probably the most intensively analysed computer simulation of letter series completion problems. Their aim was to understand how humans solved these kinds of problems and their difficulty, through the use of a computer model. Their simulation requires four basic subroutines for obtaining the correct solution: 

\begin{enumerate}
	\item  {\bf{The detection of interletter relations.}} It is assumed that the subjects have in memory the English alphabet, and the alphabet backwards, so it is also assumed that the subject have the concept of \emph{same} or \emph{equal} --- e.g., \emph{c} is the same as \emph{c}; the concept of \emph{next} on a list --- e.g., \emph{d} is next to \emph{c} on the alphabet, and \emph{f} to \emph{g} on the backward alphabet. Therefore, Thurstone series completion problems in Table \ref{fig:letter} can be solved using three interletter relations: \emph{identity}, \emph{next}, and \emph{backwards
next}.
	
	\item  {\bf{The discovery of periodicity.}} It is assumed that the subjects are able to produce a cyclical pattern (or finding regularly occurring breaks) - e.g., to cycle on the list \emph{at} in order to produce \emph{atatatat\dots}. The subject can also discover periods of letter lengths (relations occurring at a fixed interval) - e.g, having the series \emph{atbataatbat\_}, we can mark it off in periods of three letters length (\emph{ata}, \emph{atb}, \emph{ata} and \emph{at\_}). Here we observe that the first and the second position of each period are occupied, respectively, by an \emph{a} and a \emph{t} which is a symple cycle of \emph{a's} and \emph{t's} (as previously), an the third position is occupied by the cycle \emph{ba ba \dots}.

	\item  {\bf{The completion of a pattern description.}} This component involves the assembly of the two previous components in order to generate the entire series. 
	
	\item  {\bf{Extrapolation.}} Hold the pattern discovered in order to continue the generation of the letter series.
\end{enumerate}

For our purposes, only the first two subroutines will be taken into account (the last ones have to do with implementation aspects of \cite{Kotovsky1973399}).

\begin{figure}
\centering
{\sffamily\small
\begin{tabular}{ll} 
{\bf{1.}}& cdcdcdcd\_  \\
{\bf{2.}}& aaabbbcccdd\_  \\
{\bf{3.}}& atbataatbat\_  \\
{\bf{4.}}& abmcdmefmghm\_  \\
{\bf{5.}}& defgefghfghi\_  \\
{\bf{6.}}& qxapxbqxa\_  \\
{\bf{7.}}& aducuaeuabuafua\_  \\
{\bf{8.}}& mabmbcmcdm\_  \\
{\bf{9.}}& urtustuttu\_  \\
{\bf{10.}}& abyabxabwab\_  \\
{\bf{11.}}& rscdstdetuef\_  \\
{\bf{12.}}& npaoqapraqsa\_  \\
{\bf{13.}}& wxaxybyzczadab\_  \\
{\bf{14.}}& jkqrklrslmst\_  \\
{\bf{15.}}& pononmnmlmlk\_  \\

\end{tabular}
}
\caption{Letter series completion test problems from \cite{SimonKotovsky1963}.}
\label{fig:letter}
\end{figure}

\cite{Kotovsky1973399} included several pattern generator variants conceived to solve the series, which searched for patterns using IPL-V, Newell's  information processing language V \cite{newell1961}. 
All variants in \cite{SimonKotovsky1963} (that became progressively more powerful from A to D) are based on the same simple relation recognising processes . The variants show different degrees of success in describing the 15 test sequences (in Figure \ref{fig:letter}) and have been compared with different sets of subjects. The goal of writing different variants was to establish a ranking between ``hard" problems and ``easy" ones. The most successful one (variant D) solves 13 of the 15 test, outperforming 10 of the 12 subjects \cite[Table 3]{SimonKotovsky1963}.

\comment{
\begin{figure}
	\centering
		\includegraphics[width=0.5\textwidth]{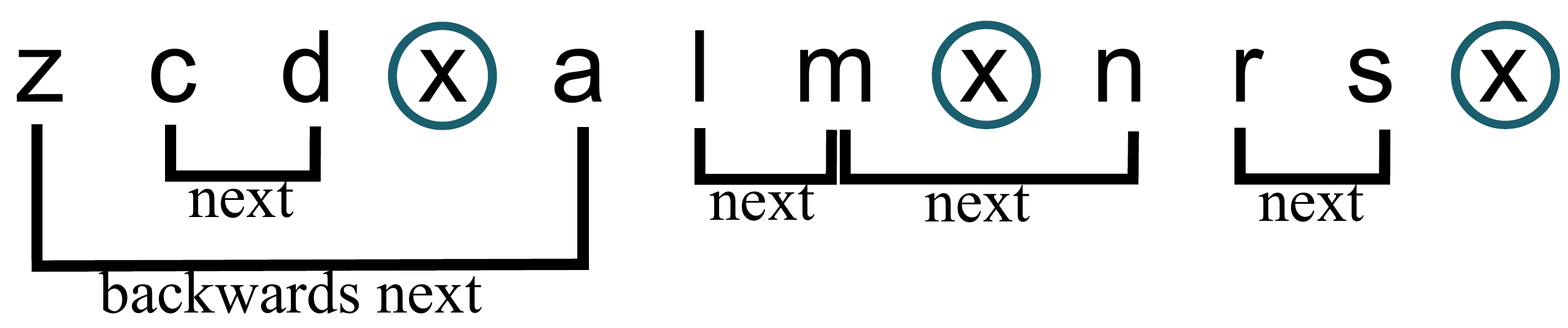}
	\caption{Interletter Relations (black underlines) and Periodicity Patterns (blue circles)}
	\label{fig:letter2}
\end{figure}
}

\subsubsection{Using \gerl}

The first step to deal with Thurstone letter series problems is to code the examples as a equations in order to be correctly addressed by \gerl. The letter series ($lhs$ of the examples) will be coded as lists of characters (or strings). The $rhs$ will be the character following the input letter series. Below we can see an example of an instance:

$$ e_1:  thurstone(``cdcdcdcd") \rightarrow ``c" $$

Since each letter series is a problem itself, we need to provide the system with more than one train instance as we have done for addressing the problem described in the previous section. We do that by decomposing the input letter series in each example into series of increasing length. For instance, from the previous example $e_1$ we create the following training instances:

\begin{equation*}
e_1:  thurstone(``cdcdcdcd") \rightarrow ``c"
  \left\lbrace
  \begin{array}{l}
    {e_1}_{.1}:  thurstone(``cd") \rightarrow ``c"\\
    {e_1}_{.2}:  thurstone(``cdc") \rightarrow ``d" \\
		{e_1}_{.3}:  thurstone(``cdcd") \rightarrow ``c" \\
		{e_1}_{.4}:  thurstone(``cdcdc") \rightarrow ``d" \\
		{e_1}_{.5}:  thurstone(``cdcdcd") \rightarrow ``c" \\
		{e_1}_{.6}:  thurstone(``cdcdcdc") \rightarrow ``d" \\
		{e_1}_{.7}:  thurstone(``cdcdcdcd") \rightarrow ``c"
  \end{array}
  \right.
\end{equation*}

Following the previous ideas about the basic subroutines needed to generate the correct solution in the Thurstone letter series \cite{SimonKotovsky1963,Kotovsky1973399}, it is easy to see that we need some operators in order to (a) work with strings, (b) establish interletter relations and, finally, (c) deal with the periodicities.  

For the former objective (a) we use $meta\_replace(Rt_1,StrucFunc(R_1))$ and $meta\_replace(Rt_1,StrucFunc(L_1))$ to replace the $rhs$ of the rules by an application of one built-in-function that works with lists ($head$, $tail$, $last$ or $init$). We will have as many operators as functions can be placed at the generic parameter $StrucFunc$. We can observe that $StrucFunc$ can be applied to the $L_1$ or $Rt_1$ position, depending on, respectively, if we want to work with the input string or with the result of the previous application of (different) operators. 

For the second objective (b), we can also use the meta-operator $meta\_replace$, instantiated as $replace(Rt_1,RelationFunc(R_1))$, where $RelationFunc$ is an interletter function, $previous$ or $next$, 
coded in the Background Knowledge. 

Finally, in order to deal with the periodicities (c), we use $insert$ to insert conditions about the position of the following letter (length of the input letter plus one) which is the same as finding regularly occurring breaks in a given relation. 
The aim is to allow \gerl to learn problems with more than one pattern (for instance ``abxcdx" where the following letter is ``x" if the position of the missing letter is multiple of 3, and the $next$ of the last letter, otherwise). In short, since the most common fixed intervals  (observed at the examples) are periods of two or three lengths, this operator has to insert, as rule conditions (at position $G_1$), questions about the position of the following letter ($position(List,Interval) = (length(List) + 1) \; mod \; Interval = 0 \;|\; Interval \in \{2,3\}$)  and, also, for the opposite condition ($\neg position(List,Interval)=(length(List)+1) \; mod \; Interval \ne 0 \;|\; Interval \in \{2,3,4\})$. 
The function $position$ is defined in the background knowledge. The meta-operator $meta\_insert$ will be instantiated as $meta\_insert(G_1, position(L_1,Interval))$ or $meta\_insert(G_1, \neg position(L_1,Interval))$ generating as many operators as intervals we have.

As usually in other problems, we need also a generalisation operator which will be applied over the input attribute to generalise it in order to try to generate a more general rule ($meta\_replace(L_1,V_{String})$).

In order to clarify the operation of the above operators, next we show an example of resolution of the letter series problem  ${e_1}_{.3}$:

{\small

\[\begin{array}{rcl} 

\blue[op_1](thurstone(``cdcd") \rightarrow ``c") & \Rightarrow & thurstone(V_{String}) \rightarrow ``c" \\ 
\blue[op_2](thurstone(V_{String}) \rightarrow ``c") & \Rightarrow & thurstone(V_{String}) \rightarrow init(V_{String}) \\ 
\blue[op_3](thurstone(V_{String}) \rightarrow init(V)) & \Rightarrow & thurstone(V_{String}) \rightarrow last(init(V_{String}))  

\end{array}\]
}

\noindent where

\[\begin{array}{ll} 

\blue[op_1]\equiv & meta\_replace(L_1, V_{String})\\ 
\blue[op_2]\equiv & meta\_replace(Rt_1,init(L_1)) 	\\ 
\blue[op_3]\equiv & meta\_replace(Rt_1,last(R_1)) 	\\

\end{array}\]

It is easy to see that the example ${e_1}_{.3}$ follows a regular pattern just alternating the characters ``c" and ``d", so the last rule obtained $thurstone(V) \rightarrow last(init(V))$ returns the right solution whatever the input (covers all seven instances generated from example ${e_1}_{.3}$). For instance, if $V_{String}=``cdcdc"$, then $init(V)=``cdcd"$ and $last(init(V)=``d"$, which is the correct following letter and is a general solution for all letter series which follow the same pattern as the previous example.

\subsubsection{Results}

\gerl has been tested on 15 problems of the Thurstone Letter Series Completion \ref{fig:letter} from \cite{SimonKotovsky1963}. As we had said previously, in \cite{SimonKotovsky1963} several variants of a pattern generator have been written where one of them (variant D) was able to score better than 10 of 12 human subjects on 15 problems of the Thurstone Letter Series Completion \cite[Table 3]{SimonKotovsky1963}. 

 With the operators provided, the system learns 14 of the 15 test sequences outperforming all four variants (A to D from \cite{SimonKotovsky1963}) and 11 of the 12 human subjects which were tested in the same work. The learned solutions are shown in Table\ref{fig:resletter}.

\begin{table}
{\scriptsize{

\centering
{\sffamily\tiny
\begin{tabular}{clllccc}
\hline 
Id														& Problem														& Solution																																						& Steps 			& $E^+$				& $o$	\\
\hline

{\bf{1.}} 										& cdcdcdcd\_  											& $thurstone(V) \rightarrow last(init(V))	$																						&  42					& 5						  &	3   \\
{\bf{2.}} 										& aaabbbcccdd\_  										& $thurstone(V) \rightarrow next(init(init(V)))$																			&  91     		&	9							&	4  	\\
{\bf{3.}} 										& atbataatbat\_  										& $thurstone(V) \rightarrow last(init(init(init(init(init(V))))))$										&	 131				&	7							& 7		\\	
\multirow{2}{*}{{\bf{4.}}} 		& \multirow{2}{*}{abmcdmefmghm\_} 	&	$thurstone(V)$ when $length(V)$ mod $3 = 0$ $\rightarrow last(init(init(V)))$				&	 174				&	8							& 5		\\
															&																		&	$thurstone(V)$ when $length(V)$ mod $3 \ne 0$ $\rightarrow next(init(V))$						&							&								& 4		\\
{\bf{5.}} 										& defgefghfghi\_  									&	$thurstone(V) \rightarrow next(init(init(init(V))))$																&	 105				&	8							& 5		\\
{\bf{6.}} 										& qxapxbqxa\_  											& $thurstone(V) \rightarrow last(init(init(init(init(init(V))))))$          					&	 129				&	7							& 7		\\
{\bf{7.}} 										& aducuaeuabuafua\_ 								&																																											&							&								& 11	\\
\multirow{2}{*}{{\bf{8.}}} 		& \multirow{2}{*}{mabmbcmcdm\_}  		& $thurstone(V)$ when $length(V)$ mod $3 = 0$ $\rightarrow last(init(init(V)))$				&	 165				&	8							& 5		\\
															&																		&	$thurstone(V)$ when $length(V)$ mod $3 \ne 0$ $\rightarrow next(init(init(V)))$			&							&								& 5		\\
\multirow{2}{*}{{\bf{9.}}}		& \multirow{2}{*}{turtustuttu\_}  	& $thurstone(V)$ when $length(V)$ mod $3 = 0$ $\rightarrow next(init(init(V)))$				&	 192 				&	9							& 5		\\
															&																		&	$thurstone(V)$ when $length(V)$ mod $3 \ne 0$ $\rightarrow last(init(init(V)))$			&							&								& 5		\\
\multirow{2}{*}{{\bf{10.}}}		& \multirow{2}{*}{abyabxabwab\_}  	& $thurstone(V)$ when $length(V)$ mod $3 = 0$ $\rightarrow previous(init(init(V)))$		&	 182				&	9							& 5		\\
															&																		&	$thurstone(V)$ when $length(V)$ mod $3 \ne 0$ $\rightarrow last(init(init(V)))$			&							&								& 5		\\
{\bf{11.}}										& rscdstdetuef\_  									&	$thurstone(V) \rightarrow next(init(init(init(init(V)))))$													&	 154				&	9							& 6		\\
\multirow{2}{*}{{\bf{12.}}}		& \multirow{2}{*}{npaoqapraqsa\_}  	& $thurstone(V)$ when $length(V)$ mod $3 = 0$ $\rightarrow last(init(init(V)))$				&	 141   			&	8							& 5		\\
															&																		&	$thurstone(V)$ when $length(V)$ mod $3 \ne 0$ $\rightarrow next(init(init(V)))$			&							&								& 5		\\
{\bf{13.}}										& wxaxybyzczadab\_  								& $thurstone(V) \rightarrow next(init(init(V)))$																			&	 99				  &	9							& 4		\\
{\bf{14.}}										& jkqrklrslmst\_ 	 									& $thurstone(V) \rightarrow next(init(init(init(V))))$ 																&	 103				&	9							& 5		\\
{\bf{15.}}										& pononmnmlmlk\_  									&	$thurstone(V) \rightarrow previous(init(init(V)))$																	&	 112				&	9							& 5		\\

\hline

\end{tabular}
}
\caption{Solution returned and steps needed by \gerl to learn the series completion test problems from \cite{SimonKotovsky1963}. The solutions contains all the information contained in sequences from which they were derived and they can be used to extrapolate the sequences indefinitely.}
\label{fig:resletter}
}}
\end{table}

\section{Discussion}\label{discussion}

In this section, we will discuss on the usefulness of \gerl and its implications, especially in terms of the complexity of the problems solved, the operators used and the number of steps needed to learn.

First of all, we will focus on how the knowledge learned by \gerl can be reused in order to decrease training between related tasks and, also, in order to accelerate learning between totally different tasks. Although the transfer of learning in Reinforcement Learning has made significant progress in recent years, there is a poor understanding about the reuse of knowledge for solving future related or different problems. In order to go beyond the previous approaches, \gerl uses an abstract representation of states and actions which facilitates the transfer of knowledge, where this abstract representation is beneficial even when the problems are not related and gives support to the idea of a general system that can perform better as it sees more and more problems. To demonstrate the usefulness of the policy reuse strategy, an example is shown is section \ref{TLexperiments}, where the ultimate goal is to use old policy information to speed up the learning process of another different problem (which a different subset of operators).

Regarding the IQ tests solved, taken into account that it has been argued that this sort of tests are the right tool to evaluate AI systems \cite{Detterman2011} and totally the opposite \cite{iq,IQnotformachines,hernandez2013universal}, in this paper we will not go into this discussion.

Unlike the other computer models solving IQ tests viewed in section \ref{IQ}, \gerl is the first system able to deal with more than one IQ test achieving equivalent learning results to the state of art and, also, compared to humans. This wide range of IQ tests addressed results in that \gerl requires certain preprocessing of the problems (input), a suitable representation and, sometimes, a large background knowledge.  However, the main difference between \gerl and the other computer models is that the former learns to solve IQ problems.


\comment{
\begin{table}[htdp]
\begin{center}
{\small
\begin{tabular}{lccccccc}\hline 
model                                                                        & years   & IQ tests           & anth.           & abs.                & learns       & raw             & results      \\ \hline 

Simon \& Kotovsky \cite{SimonKotovsky1963}                                   & 1963    & letter series      & $\times$        & $\checkmark$        & $\times$     & $\checkmark$    & human avg.   \\ 
Carpenter \cite{carpenter1990one}                                            & 1990    & Raven's PMs        & $\times$        & $\times$     				& $\times$     & $\times$ & human avg.   \\
Lovett et al. \cite{lovett2007analogy,lovett2009solving,lovett2010structure} & 2007-10 & Raven's PMs        & $\times$        & $\checkmark$        & $\times$     & $\checkmark$    & above human  \\
Lovett et al. \cite{lovett2008computational}                                 & 2008    & odd-one-out        & $\times$        & $\checkmark$        & $\times$     & $\checkmark$    & children     \\
Klenk et al. \comment{(Ext'd. Companion)} \cite{klenk2011using}              & 2011    & Bennett's Mech.    & $\times$        & $\checkmark$        & $\times$     & $\checkmark$    & not compared \\ 
McGreggor \& Goel  \cite{McGreggor2011}                                      & 2011    & odd-one-out        & $\times$        & $\checkmark$        & $\times$     & $\checkmark$    & not compared \\
Ruiz \cite{ruiz2011building}                                                 & 2011    & odd-one-out        & $\times$        & $\checkmark$        & $\times$     & $\times$        & human avg.   \\
Ragni \& Neubert \cite{ragni2012solving}                                     & 2012    & Raven's PMs        & $\checkmark$   & $\times$            	& $\times$     & $\times$        & human avg.   \\  
\gerl																																				 & 2013		 & several					  & $\times$				& $\checkmark$				& $\checkmark$ & $\times$ 			 & above human \\	

\hline\end{tabular}
}
\end{center}
\caption{Taxonomy of computer models solving IQ tests. All systems are automated. }
\label{tab:models1}
\end{table}%

}

Many of the tests seen in the previous section have a visual component (Raven's matrices and odd-one-out). While this is relevant to show how pattern recognition and image processing integrates with reasoning processes, one may wonder why we do not use IQ tests for blind people instead. This is similar to the different approaches to the odd-one-out problem used by McGreggor et al. \cite{McGreggor2011} and Ruiz \cite{ruiz2011building}. While McGreggor et al. (proudly) relies on the visual representation, Ruiz assumes that some previous feature transformation has been made. In this regard, what was considered as an `extra' (being able to process the raw information) is not seem as so in terms of the underlying abilities we want to measure, as visual recognition is usually similar for human subjects scoring very differently on these problems. In fact, CAPTCHAs (Completely Automated Public Turing test to tell Computers and Humans Apart) \cite{von2004telling}  are typically based on these pattern recognition abilities, which are not very correlated to intelligence in humans. One wonders whether tests for blind people (or people with other perceptual disabilities) would be more appropriate.

In order to solve odd-one-out problems, \gerl has made use of an abstract (R-ASCM) representation to code the examples that, with the kind of the operators implemented, provide a precise way to learn this sort of problems. In this case, the difficulty in learning the problem by \gerl radiates in the number of examples provided (the more examples provided (initial set of rules $R$), the more possible actions to apply), and in how many operators have to be applied correctly.


Regarding the way to solve Ravens's PMs, we use a feature-based coding used to represent cells, rows, and matrices: Every figure inside a cell is abstractly represented as a tuple of features: $\left\langle shape, size, quantity, position, type \right\rangle$. Five elementary relations were implemented in the background knowledge (to be used by the operators defined): Identity, One of Each, Numeric Progression, OR, and XOR. This small repertoire of relations turned out to be enough for outperforming on an IQ 100 level on the SPM test. With this abstraction and the relations, how difficult is a problem (number of steps needed to learn it) depends in how many operators have to be applied, and in the number of positive examples provided (the number of negative examples remain constant) as we can see in Table \ref{tab:resSPM}. A multiple correlation coefficient of $0.907$ confirms a strong linear association between these two variables and the number of steps.

Overall, our model matched the performance of above-average American adults on the Standard Progressive Matrices. Although the test is designed to steadily increase in difficulty throughout each section \cite{raven1996manual}, our system’s performance only depends on (the number of examples provided and on) the number operators that have to be correctly applied in order to obtain a complete and consistent solution. The only problem where \gerl fails is explained because does not follow any of the five relations described (Identity, One of Each, Numeric Progression, OR, and XOR) between the objects in a row or column.

Finally, in the previous work dealing with Thurstone letter series \cite{SimonKotovsky1963}, different degrees of success have been shown in describing the 15 test sequences depending on the difficulty of the problems. In this work the problems are classified into two groups by the length of the descriptions (and, somehow, by the length of the periods discovered): the easy ones (with simple descriptions/solutions) and the difficult ones (with simple descriptions/solutions).  As we can see in Figure \ref{fig:resletter}, \gerl also distinguishes the different problems into more or less difficult according to the number of steps needed to learn them, but, it is also related with the number of operators that have to be correctly applied and the number of positive instances provided for each letter series to solve as in the previous problems (multiple correlation coefficient of $0.944$).

\section{Conclusions and future work}\label{conclusions}

The increasing interest in learning from complex data has led to a more integrated view of this area, where the same (or similar) techniques are used for a wide range of problems using different data and pattern representations. This general view has not been accompanied by general systems that address a large variety of problems representations. Most systems need to be modified (or became completely useless) when the original data representation and structure change. Some other approaches require the data to be flattened, sliced or migrated to hyperspaces of unintelligible fictitious features. 
In fact, the most general approach can still be found in ILP (or the more general area of inductive programming). However, each system is still specific to a set of embedded operators and heuristics.

In this paper, we have proposed that more general systems can be constructed by not only giving power to data and background knowledge representation but also to a flexible operator redefinition and the reuse of heuristics across problems and systems. This flexibility also carries a computational cost. In order to address this issue we rely on two (compatible) mechanisms: the definition of customised operators, depending on the data structures and problem at hand, done by the user, using  a language for expressing operators. The other mechanisms is the use of generalised heuristics, since the use of different operators precludes the system from using specialised heuristics for each of them. The choice of the right pair of operator and rule has been reframed as a decision process, as a {\em reinforcement learning} problem.

The definition of operators is a difficult issue, and it requires some expertise and knowledge of the functional language used to express them. However, the definition of heuristics is a more difficult issue that cannot be assigned to users. Therefore, not only is this a novel approach but also allow us to better understand the role of operators and heuristics in machine learning.

We have included some illustrative examples with our latest system version, and we have seen where the flexibility stands out, since \gerl is able to solve a wide range of problems (from recursive ones to several IQ tests). Our work has several limitations. Constructing new operators is not always easy and requires expert knowledge. Also, the system becomes very inefficient when the number of examples become large.

Some possible future work is focused on using a subset of the examples as base examples and the rest for calculating coverage. We would like to include features for the operators as well.

Overall, we are conscious that our approach entails some risks, since a general system which can be instantiated to behave virtually like any other system by a proper choice of operators is an ambitious goal. We think that for complex problems that cannot be solved by the system with its predefined operators, the system can be used to investigate which operators are more suitable. In more general terms, this can be used as a system testbed, where we can learn and discover some new properties, limitations and principles for more general machine learning systems that can be used in the future.

\begin{appendices}

\section{Algorithm}

In this appendix we include a formalism of the main algorithm in \gerl. Before introducing the algorithm we require some notation. Namely: $n_v, n_c, n_f, n_s$ are, respectively, the number of variables, constants, functors and complex structures of a rule $\rho$; $MsgLen$ refers the equation \ref{msglen}; $Opt$ refers to the equation \ref{Opt}, $isRec$ checks if a rule is recursive or not, $actions$ returns the list of all possible actions and, finally, $regression\_model$ trains a regression model with the information stored in Q.

Regarding the functions in the algorithm, $initialise_R$ (Algorithm \ref{alg:initialiseR}) is in charge of initialising the set of rules $R$ with the set of positive examples provided to the system. This function makes use of the function $abstract$ (Algorithm \ref{alg:abstract}) that abstracts the description of a rule as a tuple of features (as explained in section \ref{states}). 

Moreover, the function $state$ (Algorithm \ref{alg:state}) returns abstract representation (as a tuple of features) of the actual state of the system (also explained in section \ref{states}). $initialise_Q$ (Algorithm \ref{alg:initialiseQ}) initialises the table Q  with all possible combination of actions (rules $\times$ operators) with $q$ values equal to the input parameter $q$. Finally, $stopCriterion$ (Algorithm \ref{alg:stop}) is the function in charge of checking if the \emph{stop criterion} have been fulfilled.

The function \gerl (Algorithm \ref{alg:gerl}) is the main procedure of our system which, after initialising the set of rules $R$, programs $P$ and the table $Q$, trains a regression model $Q_M$ in order to obtain the best action to perform, generating new rules and programs (which are added to their respective sets) until the stop criterion is reached.

\begin{algorithm}
\begin{algorithmic}[1]
\REQUIRE $E$ is a set of examples.
\ENSURE $\dot{R}$

\FORALL {$e \in E$}
\STATE $R \leftarrow R \cup abstract(e)$
\ENDFOR

\end{algorithmic}
\caption{$initialise_R(E)$}
\label{alg:initialiseR}
\end{algorithm}

\begin{algorithm}
\begin{algorithmic}[1]
\REQUIRE $\rho$ is functional rule, $E^+$ is a set of positive examples, $E^-$ is a set of negative examples and $K$ is the background knowledge,
\ENSURE $\dot{\rho}$

\STATE $\dot{\rho}_{\varphi_{1}} \leftarrow MsgLen(\rho)$
\STATE $\dot{\rho}_{\varphi_{2}} \leftarrow Card(\{e\in E^+: \rho \cup K \cup E^+ - e \models e  \})$
\STATE $\dot{\rho}_{\varphi_{3}} \leftarrow Card(\{e\in E^-: \rho \cup K \cup E^+ - e \models e  \})$
\STATE $\dot{\rho}_{\varphi_{4}} \leftarrow n_v(\rho)$
\STATE $\dot{\rho}_{\varphi_{5}} \leftarrow n_c(\rho)$
\STATE $\dot{\rho}_{\varphi_{6}} \leftarrow n_f(\rho)$
\STATE $\dot{\rho}_{\varphi_{7}} \leftarrow n_s(\rho)$
\STATE $\dot{\rho}_{\varphi_{8}} \leftarrow isRec(\rho)$
\STATE $\dot{\rho}_{Opt} \leftarrow Opt(\rho', E^+, E^-,K)$

\end{algorithmic}
\caption{$abstract(\rho,E^+,E^-,K)$}
\label{alg:abstract}
\end{algorithm}

\begin{algorithm}
\begin{algorithmic}[1]
\REQUIRE $R$ is a set of rules and $P$ is a set of programs.
\ENSURE $\dot{s}$

\STATE $\dot{s}_{\phi_{1}} \leftarrow \sfrac{1}{Card(P)}  \sum_{p \in P} Opt(p) $
\STATE $\dot{s}_{\phi_{2}} \leftarrow \sfrac{1}{Card(P)}  \sum_{\rho \in R} \dot{\rho}_{\varphi_{1}} $
\STATE $\dot{s}_{\phi_{3}} \leftarrow \sfrac{1}{Card(P)}  \sum_{p \in P} card(p) $

\end{algorithmic}
\caption{$state(R,P)$}
\label{alg:state}
\end{algorithm}

\begin{algorithm}
\begin{algorithmic}[1]
\REQUIRE $s$ is a state in its abstract representation, $R$ is a set of rules,  $O$ is a set of operators and $q$ is the initial q-value used to initialise the table Q.
\ENSURE $Q$
\FORALL {$o \in O$}
\FORALL {$\dot{\rho} \in R$}
\STATE $Q[\dot{s},<o,\dot{\rho}>] \leftarrow q$
\ENDFOR
\ENDFOR

\end{algorithmic}
\caption{$initialise_Q(\dot{s},R,O,q)$}
\label{alg:initialiseQ}
\end{algorithm}

\begin{algorithm}
\begin{algorithmic}[1]
\REQUIRE $P$ is a set of programs, $t$ is the current number of steps, $n$ is the number of programs last generated whose optimality value will be used in order to calculate the standard deviation $\sigma$ (to analyse the difference between them), and $\epsilon$ is the threshold used to determinate that a better program are not likely to found.

\IF{$t = \tau$}
\STATE $true$
	\ELSE \IF {$t \geq n$}
		\STATE $\bar{x} \leftarrow \frac{\sum_{i=0}^{n-1}{Opt(p_{t-i})| p\in P}}{n}$ \hspace*{6em}
		\rlap{\smash{$\left.\begin{array}{@{}c@{}}\\{}\\{}\end{array}\color{black}\right\}%
          \color{black}\begin{tabular}{l}Standard deviation. \end{tabular}$}}
		\STATE $\sigma \leftarrow \sqrt{ \frac{1}{n-1} \sum_{i=0}^{n-1}{(p_{(t-i)}-\bar{x})^2| p\in P}}$
		\IF{$\sigma \leq \epsilon$}
			\STATE $true$
		\ELSE
			\STATE $false$
		\ENDIF
	\ENDIF
\ELSE 
\STATE $false$
\ENDIF

\end{algorithmic}
\caption{$stopCriterion(P,t,\epsilon,n)$}
\label{alg:stop}
\end{algorithm}

\begin{algorithm}
\begin{algorithmic}[1]
\REQUIRE $E^+$ is a set of positive examples, $E^-$ is a set of negative examples, $K$ is the background knowledge, $O$ is a set of operators, $\tau$ is a numeric value which indicates the training periodicity of the data mining model, $q$ is the initial q-value in order to initialise the table and $\epsilon$ and $n$ are the parameters used in the $stopCriterion$ function.
\ENSURE Set of rules $R$ and programs $P$ sorted by optimality.
\STATE $R \leftarrow initialise_R(E^+)$
\STATE $P \leftarrow  R$
\STATE $\dot{s_0} \leftarrow state(R,P)$
\STATE $Q \leftarrow  initialise_Q(\dot{s}_0,R,O,q)$
\STATE $Q_M \leftarrow regression\_model(Q)$
\STATE $t \leftarrow 1$
\STATE $\dot{s_1} \leftarrow state(R,P)$
\WHILE {$\neg StopCriterion(P,t,\epsilon,n)$} 

\IF {$t \bmod \tau = 1 $} 
\STATE $Q_M \leftarrow regression\_model(Q)$
\ENDIF
\STATE $\dot{a} \leftarrow actions(O,R)$
\STATE $\dot{a}_t \leftarrow \left\langle o, \dot{\rho} \right\rangle \leftarrow \arg\max_{a \in {\cal{A}}}\left\{Q_M(\dot{s}_t,\dot{a}) \right\}$

\STATE $\rho' \leftarrow  o(\rho)$ 
\STATE $\dot{\rho}' \leftarrow abstract(\rho')$   \hspace*{2em}
		\rlap{\smash{$\left.\begin{array}{@{}c@{}}\\{}\end{array}\color{black}\right\}%
          \color{black}\begin{tabular}{l}Rule Generator.\end{tabular}$}}
										
\IF {$\dot{\rho}' \ne \dot{\rho}$}
\STATE $R \leftarrow  R \cup \dot{\rho}'$
\STATE $p_1 \leftarrow \arg\max_{p \in \left\{p_i \cup p_j | p_i,p_j \in P\right\}} Opt(p) $
\STATE $p_2 \leftarrow \arg\max_{p \in \left\{\dot{\rho}' \cup p_i | p_i \in P\right\}} Opt(p)$ \hspace*{2em}
		\rlap{\smash{$\left.\begin{array}{@{}c@{}}\\{}\\{}\end{array}\color{black}\right\}%
          \color{black}\begin{tabular}{l}Program Generator.\end{tabular}$}}
\STATE $p' \leftarrow \arg\max_{p \in\left\{p_1, p_2, \dot{\rho}'\right\}} Opt(p)$ 
\STATE $P \leftarrow  P \cup p'$
\ENDIF
\STATE $\dot{s}_{t+1} \leftarrow state(R,P)$ 
\STATE $Q[\dot{s}_{t}, \dot{a}_{t} ] \leftarrow \alpha \left[ \dot{\rho}'_{Opt} + \gamma \max_{a \in {\cal{A}}} Q_{M}(\dot{s}_{t+1},\dot{a}) \right] + (1-\alpha) Q[\dot{s}_{t},\dot{a}_{t}]$
\STATE $t \leftarrow t+1$

\ENDWHILE

\end{algorithmic}
\caption{$gErl(E^+,E^-,K,O,q,\tau,\epsilon,n)$}
\label{alg:gerl}
\end{algorithm}

\end{appendices}

\bibliography{biblio}

\begin{thebibliography}{10}

\bibitem{RavenWeb}
http://www.raventest.net, 2001.

\bibitem{citeulike:3371219}
John~R. Anderson.
\newblock {\em {Using Brain Imaging to Guide the Development of a Cognitive
  Architecture}}, pages 49--62.
\newblock Oxford University Press, New York, NY, 2007.

\bibitem{Armstrong07}
J.~Armstrong.
\newblock A history of erlang.
\newblock In {\em Proceedings of the third ACM SIGPLAN conf. on History of
  programming languages}, HOPL III, pages 1--26. ACM, 2007.

\bibitem{springerlink:Metalearning}
P.~Brazdil and Giraud-Carrier.
\newblock Metalearning: Concepts and systems.
\newblock In {\em Metalearning}, Cognitive Technologies, pages 1--10. Springer
  Berlin Heidelberg, 2009.

\bibitem{carpenter1990one}
Patricia~A Carpenter, Marcel~Adam Just, and Peter Shell.
\newblock What one intelligence test measures: A theoretical account of the
  processing in the raven progressive matrices test.
\newblock {\em Psychological review}, 97:404--431, 1990.

\bibitem{Chater03simplicity:a}
Nick Chater, Paul Vit\'{a}nyi, and Coventry~Cv Al.
\newblock Simplicity: A unifying principle in cognitive science?
\newblock In {\em Trends in Cognitive Sciences}, pages 7--19, 2003.

\bibitem{daume09searn}
H.~{Daum\'e III} and J.~Langford.
\newblock Search-based structured prediction.
\newblock 2009.

\bibitem{Detterman2011}
D.~K. Detterman.
\newblock A challenge to {W}atson.
\newblock {\em Intelligence}, 39(2-3):77 -- 78, 2011.

\bibitem{springerlink:10.1007/s10994-008-5079-1}
T.~Dietterich, P.~Domingos, L.~Getoor, S.~Muggleton, and P.~Tadepalli.
\newblock Structured {M}achine {L}earning: the next ten years.
\newblock {\em Machine Learning}, 73:3--23, 2008.

\bibitem{IQnotformachines}
D.~L. Dowe and J.~Hern{\'a}ndez-Orallo.
\newblock {IQ} tests are not for machines, yet.
\newblock {\em Intelligence}, 40(2):77--81, 2012.

\bibitem{KDriessensThesis}
K~Driessens.
\newblock {\em Relational Reinforcement Learning. Phd Thesis}.
\newblock Department of Computer Science, K.U.Leuven, Leuven, Belgium, 2004.

\bibitem{Dzeroski:2006:TGF:1777194.1777213}
S.~D\v{z}eroski.
\newblock Towards a general framework for data mining.
\newblock KDID'06, pages 259--300, Berlin, Heidelberg, 2007. Springer-Verlag.

\bibitem{1007694015589}
S.~D\v{z}eroski, L.~De~Raedt, and K.~Driessens.
\newblock Relational reinforcement learning.
\newblock {\em Machine Learning}, 43:7--52, 2001.
\newblock 10.1023/A:1007694015589.

\bibitem{DL01}
S.~Dzeroski and N.~Lavrac, editors.
\newblock {\em Relational Data Mining}.
\newblock Springer-Verlag, 2001.

\bibitem{vicentsim06}
V.~Estruch, C.~Ferri, J.~Hern{\'a}ndez-Orallo, and M.~J. Ram\'{\i}rez-Quintana.
\newblock Similarity functions for structured data. {An} application to
  decision trees.
\newblock {\em Inteligencia Artificial, Revista Iberoamericana de Inteligencia
  Artificial}, 10(29):109--121, 2006.

\bibitem{coin2012}
V.~Estruch, C.~Ferri, J.~Hern{\'a}ndez-Orallo, and M.J. Ram\'{\i}rez-Quintana.
\newblock {Bridging the Gap between Distance and Generalisation}.
\newblock {\em Computational Intelligence}, pages no--no, 2012.

\bibitem{evans1964program}
T.~G. Evans.
\newblock A program for the solution of a class of geometric-analogy
  intelligence-test questions.
\newblock Technical report, DTIC Document 1964, also appeared in 1968 in Minsky
  M. (ed.) Semantic Information Processing, pp. 271-353, MIT Press, Cambridge,
  Massachussets., 1964.

\bibitem{feingold1940culture}
S.N. Feingold.
\newblock {\em A Culture Free Intelligence Test Evaluation of Cultural
  Influence on Test Scores}.
\newblock Clark University, 1940.

\bibitem{Fernandez06probabilisticpolicy}
F.~Fernandez and M.~Veloso.
\newblock Probabilistic policy reuse in a {R}einforcement {L}earning agent.
\newblock In {\em AAMAS ’06}, pages 720--727. ACM Press, 2006.

\bibitem{Ferri-RamirezHR01}
C.~Ferri, J.~Hern{\'a}ndez-Orallo, and M.J. Ram\'{\i}rez-Quintana.
\newblock Incremental learning of functional logic programs.
\newblock In Herbert Kuchen and Kazunori Ueda, editors, {\em Functional and
  Logic Programming}, volume 2024 of {\em Lecture Notes in Computer Science},
  pages 233--247. Springer Berlin Heidelberg, 2001.

\bibitem{Gaertner05}
T.~G{\"a}rtner.
\newblock {\em Kernels for Structured Data}.
\newblock PhD thesis, Universitat Bonn, 2005.

\bibitem{hernandez2013universal}
Jos{\'e} Hern{\'a}ndez-Orallo, David~L Dowe, and M~Hern{\'a}ndez-Lloreda.
\newblock Universal psychometrics: Measuring cognitive abilities in the machine
  kingdom.
\newblock {\em Cognitive Systems Research}, 2013.

\bibitem{hernandez1998inverse}
Jos{\'e} Hern{\'a}ndez-Orallo and M~Jos{\'e} Ram{\'\i}rez-Quintana.
\newblock Inverse narrowing for the induction of functional logic programs.
\newblock In {\em APPIA-GULP-PRODE}, pages 379--392. Citeseer, 1998.

\bibitem{hernandez1999strong}
Jos{\'e} Hern{\'a}ndez-Orallo and M~Jos{\'e} Ram{\'\i}rez-Quintana.
\newblock A strong complete schema for inductive functional logic programming.
\newblock In {\em Inductive Logic Programming}, pages 116--127. Springer, 1999.

\bibitem{Holland00}
J.~Holland and Booker.
\newblock What is a learning classifier system?
\newblock In {\em Learning Classifier Systems}, volume 1813 of {\em LNCS},
  pages 3--32. 2000.

\bibitem{Holmes200223}
J.~H. Holmes, P.~Lanzi, and W.~Stolzmann.
\newblock Learning classifier systems: New models, successful applications.
\newblock {\em Information Processing Letters}, 2002.

\bibitem{Carroll02fixedvs}
J.Carroll.
\newblock Fixed vs {D}ynamic {S}ub-transfer in {R}einforcement {L}earning.
\newblock In {\em ICMLA'02}. CSREA Press, 2002.

\bibitem{Kitzelmann10}
E.~Kitzelmann.
\newblock Inductive programming: A survey of program synthesis techniques.
\newblock In {\em 3rd Workshop AAIP}, volume 5812 of {\em LNCS}, 2010.

\bibitem{Koller:1997:HCD:645526.657130}
D.~Koller and M.~Sahami.
\newblock Hierarchically classifying documents using very few words.
\newblock In {\em Proceedings of the Fourteenth International Conference on
  Machine Learning}, ICML '97, pages 170--178, San Francisco, CA, USA, 1997.
  Morgan Kaufmann Publishers Inc.

\bibitem{Kotovsky1973399}
Kenneth Kotovsky and Herbert~A. Simon.
\newblock Empirical tests of a theory of human acquisition of concepts for
  sequential patterns.
\newblock {\em Cognitive Psychology}, 4(3):399 -- 424, 1973.

\bibitem{Lafferty:2001:CRF:645530.655813}
J.~Lafferty and A~McCallum.
\newblock Conditional random fields: Probabilistic models for segmenting and
  labeling sequence data.
\newblock ICML '01, pages 282--289, 2001.

\bibitem{Li:2008:IKC:1478784}
Ming Li and Paul~M.B. Vit\'{a}nyi.
\newblock {\em An Introduction to Kolmogorov Complexity and Its Applications}.
\newblock Springer Publishing Company, 3 edition, 2008.

\bibitem{AAAI06-yaxin}
Y.~Liu and P.~Stone.
\newblock Value-function-based transfer for reinforcement learning using
  structure mapping.
\newblock AAAI, pages 415--20, July 2006.

\bibitem{Lloyd01knowledgerepresentation}
J.~W. Lloyd.
\newblock Knowledge representation, computation, and learning in higher-order
  logic.
\newblock 2001.

\bibitem{lovett2007analogy}
Andrew Lovett, Kenneth Forbus, and Jeffrey Usher.
\newblock Analogy with qualitative spatial representations can simulate solving
  raven’s progressive matrices.
\newblock In {\em Proceedings of the 29th Annual Conference of the Cognitive
  Society}, 2007.

\bibitem{lovett2010structure}
Andrew Lovett, Kenneth Forbus, and Jeffrey Usher.
\newblock A structure-mapping model of raven’s progressive matrices.
\newblock In {\em Proceedings of CogSci}, volume~10, pages 2761--2766, 2010.

\bibitem{lovett2008computational}
Andrew Lovett, Kate Lockwood, and Kenneth Forbus.
\newblock A computational model of the visual oddity task.
\newblock In {\em the Proceedings of the 30th Annual Conference of the
  Cognitive Science Society. Washington, DC}, 2008.

\bibitem{Maes2009SPRL}
F.~Maes, L.~Denoyer, and P.~Gallinari.
\newblock Structured prediction with reinforcement learning.
\newblock {\em Machine Learning Journal}, 77(2-3):271--301, 2009.

\bibitem{DBLP:conf/ausai/Martinez-PlumedEFHR10}
F.~Mart\'{\i}nez-Plumed, V.~Estruch, C.~Ferri, J.~Hern{\'a}ndez-Orallo, and
  M.~J. Ram\'{\i}rez-Quintana.
\newblock Newton trees.
\newblock In {\em Australasian Conference on Artificial Intelligence}, volume
  6464 of {\em LNCS}, pages 174--183, 2010.

\bibitem{FmartinezNFMCP12}
Fernando Mart\'{\i}nez-Plumed, C\`esar Ferri, Jos\'e Hern\'andez-Orallo, and
  M.Jos\'e Ramírez-Quintana.
\newblock Learning with configurable operators and rl-based heuristics.
\newblock In Annalisa Appice, Michelangelo Ceci, Corrado Loglisci, Giuseppe
  Manco, Elio Masciari, and ZbigniewW. Ras, editors, {\em New Frontiers in
  Mining Complex Patterns}, volume 7765 of {\em Lecture Notes in Computer
  Science}, pages 1--16. Springer Berlin Heidelberg, 2013.

\bibitem{mccallum2003handbook}
R.S. McCallum.
\newblock {\em Handbook of Nonverbal Assessment}.
\newblock Kluwer Academic/Plenum Publishers, 2003.

\bibitem{McGreggor2011}
Keith McGreggor and Ashok Goel.
\newblock Finding the odd one out: a fractal analogical approach.
\newblock In {\em Proceedings of the 8th ACM conference on Creativity and
  cognition}, pages 289--298, New York, NY, USA, 2011. ACM.

\bibitem{mcgreggor2010fractal}
Keith McGreggor, Maithilee Kunda, and Ashok Goel.
\newblock A fractal analogy approach to the raven’s test of intelligence.
\newblock In {\em AAAI workshops at the 24th AAAI conference on Artificial
  Intelligence}, pages 69--75, 2010.

\bibitem{Mehta05transferin}
N.~Mehta.
\newblock Transfer in variable-reward hierarchical reinforcement learning.
\newblock In {\em In Proc. of the Inductive Transfer workshop at NIPS}, 2005.

\bibitem{Mug99}
S.~H. Muggleton.
\newblock Inductive logic programming: Issues, results, and the challenge of
  learning language in logic.
\newblock {\em Artificial Intelligence}, 114(1--2):283--296, 1999.

\bibitem{mug95}
Stephen Muggleton.
\newblock Inverse entailment and progol.
\newblock {\em New Generation Computing}, 13(3-4):245--286, 1995.

\bibitem{newell1961}
Allen Newell.
\newblock {\em Information processing language-V manual}.
\newblock Prentice-Hall, 1961.

\bibitem{Plo70}
G.~Plotkin.
\newblock A note on inductive generalization.
\newblock {\em Machine Intelligence}, 5, 1970.

\bibitem{Price03acceleratingreinforcement}
B.~Price and C.~Boutilier.
\newblock Accelerating {R}einforcement {L}earning through implicit imitation.
\newblock {\em Journal of Artificial Intelligence Research}, 19:2003, 2003.

\bibitem{ragni2012solving}
Marco Ragni and Stefanie Neubert.
\newblock Solving raven’s {IQ}-tests: An ai and cognitive modeling approach.
\newblock In {\em ECAI}, pages 666--671. IOS Press, 2012.

\bibitem{raven12court}
J~Raven.
\newblock Court jh (1998) manual for raven’s progressive matrices and
  vocabulary scales.
\newblock {\em Oxford: Oxford Psychologists}, 12:G60p.

\bibitem{raven1992}
J.~C. Raven, J.~H. Court, and J.~Raven.
\newblock {\em Manual for Raven's Progressive Matrices and Vocabulary Scale}.
\newblock San Antonio, TX: Psychological Corporation, 1992.

\bibitem{raven1996manual}
J.C. Raven and J.H. Court.
\newblock {\em Manual for Raven's Progressive Matrices and Vocabulary Scales:
  Standard progressive matrices}.
\newblock Manual for Raven's Progressive Matrices and Vocabulary Scales. Oxford
  Psychologists Press, 1996.

\bibitem{ruiz2011building}
Philippe~E Ruiz.
\newblock Building and solving odd-one-out classification problems: A
  systematic approach.
\newblock {\em Intelligence}, 39(5):342--350, 2011.

\bibitem{iq}
P.~Sanghi and D.~L. Dowe.
\newblock A computer program capable of passing {I.Q.} tests.
\newblock In P.~P. Slezak, editor, {\em {P}roc. of the {J}oint {I}nternational
  {C}onference on {C}ognitive {S}cience, 4th {ICCS} {I}nternational
  {C}onference on {C}ognitive {S}cience \& 7th {ASCS} {A}ustralasian {S}ociety
  for {C}ognitive {S}cience ({ICCS}/{ASCS}-2003)}, pages 570--575, Sydney, NSW,
  Australia, 13-17 July 2003.

\bibitem{Schmidhuber:2004:OOP:969909.969942}
J.~Schmidhuber.
\newblock Optimal ordered problem solver.
\newblock {\em Maching Learning}, 54(3):211--254, March 2004.

\bibitem{SimonKotovsky1963}
H.~A. Simon and K.~Kotovsky.
\newblock Human acquisition of concepts for sequential patterns.
\newblock {\em Psychological Review}, 70(6):534, 1963.

\bibitem{aleph}
A.~Srinivasan.
\newblock {\em {The Aleph Manual (Technical Report)}}, 2004.

\bibitem{Strannegard2013progressive}
Claes Stranneg{\aa}rd, Simone Cirillo, and Victor Str{\"o}m.
\newblock An anthropomorphic method for progressive matrix problems.
\newblock {\em Cognitive Systems Research}, 22–23(0):35 -- 46, 2013.

\bibitem{sutton98a}
R.~Sutton.
\newblock {\em Reinforcement Learning: An Introduction}.
\newblock {MIT} Press, 1998.

\bibitem{Sutton99betweenmdps}
R.~Sutton.
\newblock Between {MDP}s and semi-{MDP}s: A framework for temporal abstraction
  in {R}einforcement {L}earning.
\newblock {\em Artificial Intelligence}, 112:181--211, 1999.

\bibitem{sutton1998reinforcement}
R.~S. Sutton and A.~G. Barto.
\newblock {\em {Reinforcement learning: An introduction}}.
\newblock MIT press, 1998.

\bibitem{Tadepalli04}
P.~Tadepalli, R.~Givan, and K.~Driessens.
\newblock Relational reinforcement learning: An overview.
\newblock In {\em In Proc. of the Workshop on Relational Reinforcement
  Learning}, 2004.

\bibitem{TM02}
A.~Tamaddoni-Nezhad and S.~Muggleton.
\newblock A genetic algorithms approach to {ILP}.
\newblock In {\em Proc. of the 12th Int. Conf. on Inductive logic programming},
  ILP'02, pages 285--300, Berlin, Heidelberg, 2003. Springer-Verlag.

\bibitem{JMLR09-taylor}
M.~Taylor and P.~Stone.
\newblock Transfer learning for {R}einforcement {L}earning domains: A survey.
\newblock {\em Journal of Machine Learning Research}, 10(1):1633--1685, 2009.

\bibitem{thurstone1941factorial}
L.L. Thurstone and T.G. Thurston.
\newblock {\em Factorial studies of intelligence}.
\newblock Psychometrika monograph suplements. The University of Chicago press,
  1941.

\bibitem{Tsochantaridis:2004:SVM:1015330.1015341}
I.~Tsochantaridis, T.~Hofmann, T.~Joachims, and Y.~Altun.
\newblock Support vector machine learning for interdependent and structured
  output spaces.
\newblock In {\em ICML}, 2004.

\bibitem{Armstrong:1996:CPE:229883}
Robert Virding, Claes Wikstr\"{o}m, and Mike Williams.
\newblock {\em Concurrent programming in ERLANG (2nd ed.)}.
\newblock Prentice Hall International (UK) Ltd., Hertfordshire, UK,, 1996.

\bibitem{von2004telling}
L.~{v}on Ahn, M.~Blum, and J.~Langford.
\newblock {Telling humans and computers apart automatically}.
\newblock {\em Communications of the ACM}, 47(2):56--60, 2004.

\bibitem{Wallace01081968}
C.~S. Wallace and D.~M. Boulton.
\newblock An information measure for classification.
\newblock {\em The Computer Journal}, 11(2):185--194, 1968.

\bibitem{DBLP:journals/cj/WallaceD99a}
C.~S. Wallace and D.~L. Dowe.
\newblock Refinements of {MDL} and {MML} coding.
\newblock {\em Comput. J.}, 42(4):330--337, 1999.

\bibitem{Wallace:2005:SII:1051763}
C.S. Wallace.
\newblock {\em Statistical and Inductive Inference by Minimum Message Length
  (Information Science and Statistics)}.
\newblock Springer-Verlag New York, Inc., Secaucus, NJ, USA, 2005.

\bibitem{springerlink:10.1007/BF00992698}
C.~Watkins and P.~Dayan.
\newblock Q-learning.
\newblock {\em Machine Learning}, 8:279--292, 1992.

\end{thebibliography}

\end{document}